%% file: main.tex
\documentclass{article}

\PassOptionsToPackage{numbers, compress}{natbib}

     \usepackage[preprint]{main}

\usepackage[utf8]{inputenc} 
\usepackage[T1]{fontenc}    
\usepackage{hyperref}       
\usepackage{url}            
\usepackage{booktabs}       
\usepackage{amsfonts}       
\usepackage{nicefrac}       
\usepackage[titlenumbered,ruled,linesnumbered]{algorithm2e}
\DontPrintSemicolon
\usepackage{wrapfig}

\usepackage{graphicx}
\usepackage{amsmath}
\usepackage{subcaption}
\usepackage{wrapfig}
\usepackage{color,soul}
\usepackage{enumitem}
\usepackage[dvipsnames]{xcolor}
\usepackage{amssymb}
\usepackage{wasysym}

\title{A Conceptual Framework for Lifelong Learning}

\author{%
  Charles X. Ling \\
  Department of Computer Science\\
  Western University\\
  London, ON, Canada \\
  \texttt{charles.ling@uwo.ca} \\
  \And
  Tanner Bohn \\
  Department of Computer Science \\
  Western University \\
  London, ON, Canada \\
  \texttt{tbohn@uwo.ca}
}

\begin{document}

\maketitle

\begin{abstract}

Humans can learn a variety of concepts and skills incrementally over the course of their lives while exhibiting many desirable properties, such as continual learning without forgetting, forward transfer and backward transfer of knowledge, and 
learning a new concept or task with only a few examples. 
Several lines of machine learning research, such as lifelong learning, few-shot learning, and transfer learning, attempt to capture these properties. However, most previous approaches can only demonstrate subsets of these properties, often by different complex mechanisms. In this work, we propose a simple yet powerful unified framework that supports almost all of these properties and approaches through {\em one} central mechanism. 
We also draw connections between many peculiarities of human learning (such as memory loss and ``rain man'') and our framework. While we do not present any state-of-the-art results, we hope that this conceptual framework provides a novel perspective on existing work and proposes many new research directions. 

\end{abstract}



\input{sec_1_intro.tex}

\input{sec_2_overview}

\input{sec_3_framework}
\input{sec_4_connections}
\input{sec_5_parallels}
\input{sec_6_conclusion}

\newpage

\bibliographystyle{unsrt}
\bibliography{main}



    
    
    
    


\end{document}

%% file: sec_1_intro.tex
\section{Introduction}
\label{sec:introduction}

The past decade has seen significant growth in the capabilities of artificial intelligence.
Deep learning in particular has archived great successes in medical image recognition and diagnostics \cite{litjens2017survey, shen2017deep},  tasks on natural language processing \cite{radford2019language, devlinetal2019bert}, difficult games \cite{silver2017mastering}, and even farming \cite{kamilaris2018deep}.
However, deep learning models almost always need thousands or millions of training samples to perform well.
This is in a sharp contrast with human learning, which normally learns a new concept with a small number of samples. 
Other major weaknesses in current deep learning, when compared to human learning, include difficulty in 
leveraging previous learned knowledge to better learn new ones (and vice versa), learn many tasks sequentially without forgetting previous ones, and so on. 

Several lines of research in supervised learning exist to overcome these weaknesses.
Multi-task learning \cite{caruana1997multitask} considers how to learn multiple concepts at the same time such that they help each other to be learned better.
The related field of transfer learning \cite{pan2009survey} assumes that some concepts have been previously learned and we would like to transfer their knowledge to assist learning new concepts.
Few-shot learning \cite{fei2006one} aims to learn tasks with a small number of labeled data.  
Lifelong learning \cite{thrun1998lifelong, LLL} (LLL -- also known as continual \cite{parisi2019continual} or sequential learning \cite{CatastrophicForgetting}) considers how to learn and transfer skills across long sequences of tasks.  

However, most previous approaches can only demonstrate subsets of these human-like properties, often by different complex mechanisms. 
For example, 
existing lifelong learning (LLL) techniques tend to use one or more of three types of mechanisms, each of which comes with their own drawbacks and hurdles \cite{de2019continual}.
These mechanisms are based on
replay, regularization, and dynamic architecture respectively. 
See Sec.~\ref{sec:comparison} for reviews of these mechanisms and comparison.

\begin{figure}[ht]
    \centering
    \includegraphics[width=0.75\textwidth]{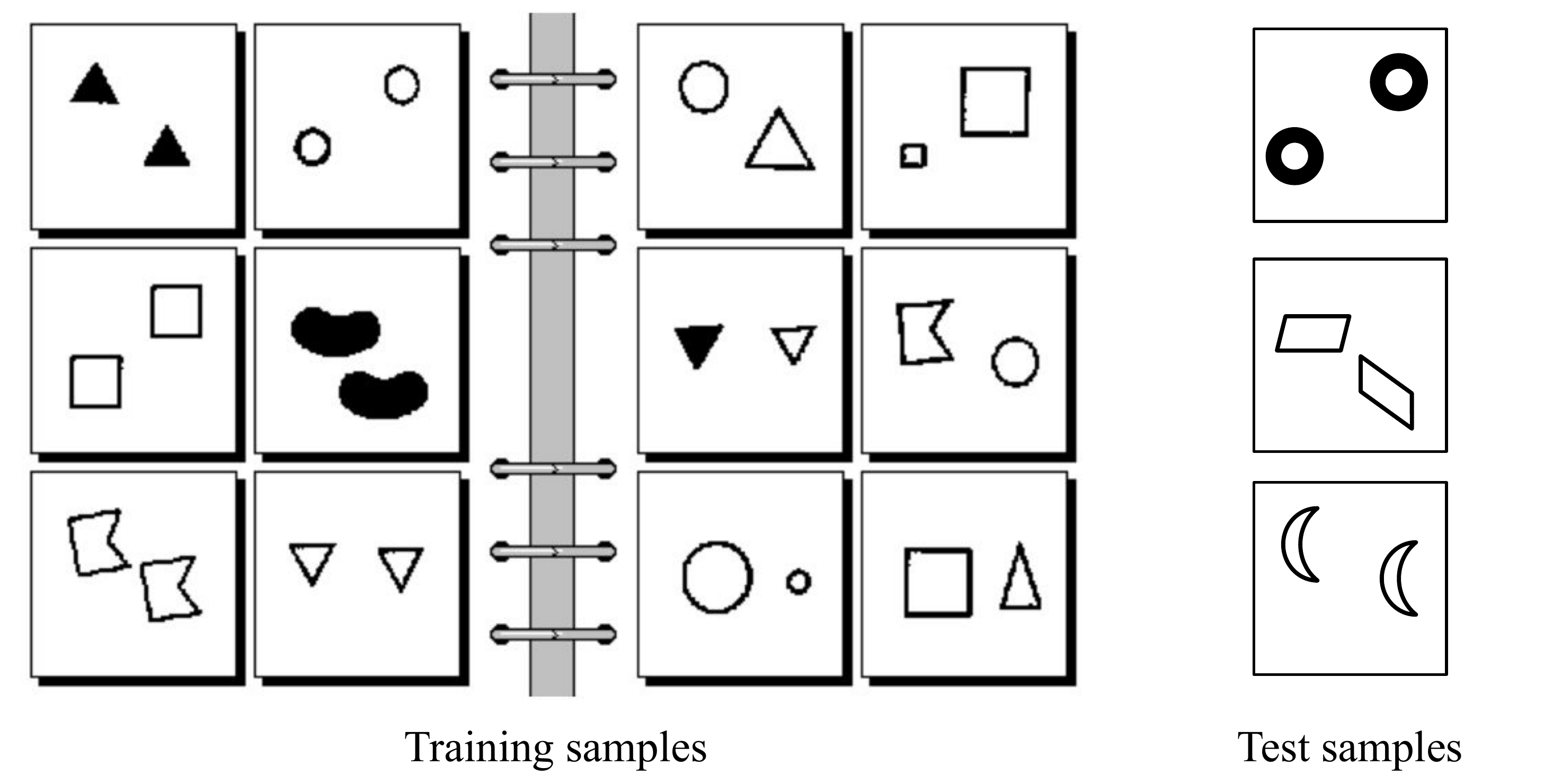}
    \caption{An example of a Bongard problem (\#57) which is relatively easy for humans to solve\protect\footnotemark{ } but especially difficult for deep learning, especially given that there are only six positive (on the left) and six negative samples (on the right). In Section~\ref{sec:rd_bongard} we discuss the application of our framework to few-shot learning to solve Bongard problems in particular.}
    \label{fig:bp57}
\end{figure}  

In terms of few-shot learning, current deep learning has difficulty achieving human-level performance, especially on visual tasks \cite{fei2006one, lake2015human}.  As an example, in Figure~\ref{fig:bp57}, given 6 images on the left belonging to one class (say positive), and 6 images on the right the other class, how would you (the reader) classify the 3 test images to the left or right?\footnotetext{Solution: on the left side, the images contain two identical shapes, but on the right side, the two shapes are different.} 
Most humans (as shown in \cite{foundalis2006phaeaco}) can identify the underlying pattern and classify correctly the test examples with only few (here 12 in total but can be fewer) samples. 
However, experiments conducted by us (unpublished) have demonstrated that the standard convolutional networks have difficulty performing better than random on a significant portion of BPs, even with millions of training images.

The problem in Figure~\ref{fig:bp57} is actually called a Bongard Problem (BP) \cite{bongard1967problem}, which is a standard supervised learning problem. There are hundreds of such problems with various difficulty levels, a majority are very difficult to learn with deep learning, especially given that there are so few samples \cite{yun2020deeper}.  For humans, most easy BPs usually are not hard with 12 (or fewer) samples. In Sec.~\ref{sec:rd_bongard} we will discuss BPs in more details and propose potential methods to solve them with a small number of examples. 


Clearly, there must be some sort of knowledge transfer from previous human concepts (such as recognizing positions, sizes, shapes) to perform well on unseen BPs.  
In addition, solving BPs will not usually make people forget how to do previous BPs or other concepts; in standard neural networks, however, catastrophic forgetting can happen. 
Even if forgetting does happen in humans, it is also ``graceful'' and gradual.  
All of these (non-forgetting, forward transfer, graceful forgetting, and so on) are properties of lifelong learning in supervised setting, to be studied in our paper.

Most previous LLL approaches can only demonstrate subsets of these human-like properties by different complex mechanisms. However, it is our belief that human lifelong concept learning with labeled examples likely uses a single (or a small set of) mechanism(s).  This is because humans learn many concepts in their lives and the process is {\em continuous without sharp boundary}. 
For example, learning a new concept (such as a new BP) or updating a learned
concept (such as receiving more data of a previously learned BP) may subtly influence other concepts learned before (and after).  Using LLL terminology, this subtle influence is due to forward- and backward-transfer, non-forgetting of previous concepts, and graceful forgetting. 
These influences appear to combine seamlessly in humans, providing effective lifelong learning capabilities. 
Thus, we wish to find a learning framework with one central mechanism that can seamlessly implement such influences\footnote{More specifically, we are aiming to describe a higher-level \textit{algorithmic} mechanism, rather than provide a model of specific neural processes associates with human learning.}, and demonstrate all (or most) human-like lifelong learning properties.

\begin{figure}[ht]
    \centering
    \includegraphics[width=0.75\textwidth]{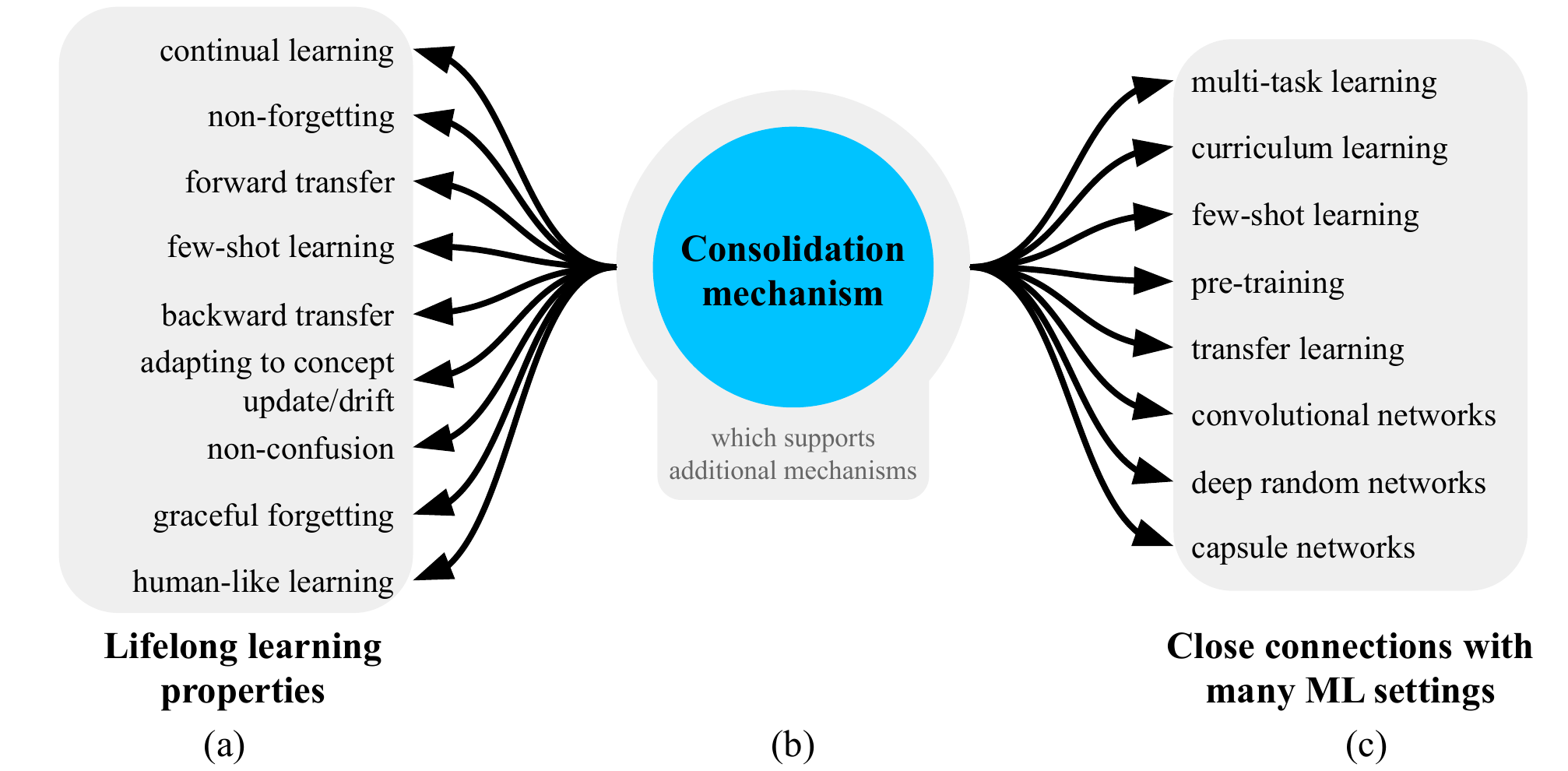}
    \caption{An overview of our unified framework: by combining a central consolidation mechanism with additional tools, we are able to exhibit many lifelong learning properties and demonstrate close connections with other machine learning settings. The set of desirable LLL properties in (a) is discussed in Sec.~\ref{sec:properties}. A description of the consolidation mechanism and its combination with the additional mechanisms shown in (b) is discussed in Sec.~\ref{sec:framework}. We similarly mark usage of the consolidation mechanism in our algorithms with a \textcolor{Cerulean}{$\CIRCLE$}. The learning settings listed in (c) where our framework can be applied is discussed in Sec.~\ref{sec:connections}. The connection between our framework and human learning in particular is expanded upon in Sec.~\ref{sec:parallels}.}
    \label{fig:framework}
\end{figure}  

In this paper, we propose such a 
\textit{unified framework with one central mechanism}. A high-level overview of the framework is depicted in Figure~\ref{fig:framework}.
We demonstrate its unified characteristic by discussing how it can illustrate many desirable LLL properties such as non-forgetting, forward and backward transfer, and graceful forgetting (Sec.~\ref{sec:framework}).
Second, we demonstrate its {\em additional} unified characteristic of being applicable to many learning settings such as LLL, multi-task learning, transfer learning, and curriculum learning (Sec.~\ref{sec:connections}). 
We also propose many potential new research topics under our framework. 

In addition, we are able to demonstrate many peculiarities seen in human learning, such as memory loss and ``rain man'', under our proposed framework (Sec.~\ref{sec:parallels}). This may shed new light on studying human learning from machine learning perspectives.


This paper may also serve as a survey for LLL and multi-task learning.    
Previous survey papers on these topics have been published. 
For example, \cite{parisi2019continual} extensively discusses both existing and emerging techniques for lifelong machine learning, draw connections and motivation from biological learning, and survey additional learning settings. In contrast, our paper discusses these topics from a unique and consistent perspective: that of being connected to a single learning mechanism.
In \cite{de2019continual}, the authors propose a framework for striking a balance between stability and plasticity to optimize non-forgetting and new task learning in fixed-size networks for LLL. However, the framework we propose can be used to demonstrate a wider array of LLL properties and applies to a variety of learning settings in a \textit{unified} way. 


This survey and conceptual framework paper is different from most deep learning papers that aim to achieve state-of-the-art results.  
Instead, our paper shows the generality of a new framework that can exhibit many properties, apply to many machine learning settings, and encompass previous works.   
In our discussions, we identify many directions to accompany previous and potential future research. 
We hope it inspires more researchers to engage in the various topics discussed in our paper.

%% file: sec_2_overview.tex
\section{Lifelong Learning and its Properties}
\label{sec:overview}

In this section we will first describe our lifelong learning (LLL) setting and its relation to other learning settings. We will then discuss a broad set of important LLL properties.  

\subsection{Lifelong Learning Setting}
\label{sec:setting}


In our lifelong setting, we mainly consider the \textit{task-incremental} classification tasks, where batches of data for new tasks arrive sequentially.  
That is, a sequence of $(T_{1}, D_{1}), (T_{2}, D_{2}), ...$ are given, where $D_{i}$ is the labeled training data of task $T_{i}$ (from the space of tasks $\mathcal{T}$), and an individual task consists of a set of classes to be learned. 
Classification models for $(T_{1}, T_{2}, ..., (T_{k}$
must be functional before $(T_{k+1}, D_{k+1})$ arrives. 
This models the incremental process of human lifelong learning. 


This LLL setting can be easily extended to other learning settings such as  multi-task learning where all $T_{i}$ and $D_{i}$ are given at the same time (also see Sec.~\ref{sec:rd_curriculum} and \ref{sec:backwardtransfer}).  It can also be extended to meta-learning (Sec.~\ref{sec:nonforgetting} and \ref{sec:forwardtransfer}) where the knowledge learned for $T_{i}$ only provides a starting point to learn $T_{j}$ more effectively. Additionally, it can be extended to curriculum learning (Sec.~\ref{sec:rd_curriculum}) where $T_{i}$ are able to be ordered from ``simple'' to ``complex'' based on a given difficulty metric.


\subsection{Lifelong Learning Properties}
\label{sec:properties}

Here we discuss at a high level several properties a LLL approach would ideally exhibit.

\textbf{Continual learning and testing}: Before starting to learn a new task $T_{j}$, a LLL approach should be able to perform well on all $T_{i<j}$. 
While learning the new task $T_{j}$,  LLL should {\em only} use data $D_{j}$. 
This is a startling contrast with the standard multi-task (batch) learning where data of all tasks are used for training at the same time.  
This ensures that the model will be consistently useful when continually learning tasks with high computational and data efficiency.  
See Sec.~\ref{sec:expansion} for details.

\textbf{Non-forgetting}: This is the ability to avoid catastrophic forgetting \cite{CatastrophicForgetting}, where learning $T_{j}$ causes a dramatic loss in performance on one or more $T_{i<j}$. Ideally, learning $T_{j}$ using only the data of $T_{j}$ 
would not affect $T_{i<j}$.
Due to the tendency towards catastrophic forgetting, non-lifelong learning approaches would require retraining on data for all tasks together to avoid forgetting. 
This may reduce computational and data efficiency. 
A property opposite to non-forgetting is \textbf{graceful forgetting} \cite{aljundi2018memory}, as often seen in humans. Learning new tasks may require additional model capacity, and when this is not possible, the model can perform graceful forgetting of unimportant tasks to free up capacity for new tasks. See Sec.~\ref{sec:nonforgetting} and \ref{sec:gracefulforgetting} for details.

\textbf{Forward transfer}: This is the ability to learn new tasks, $T_{\geq i}$, easier and better following earlier learned tasks $T_{<i}$, also known as knowledge transfer \cite{pan2009survey}.

\textbf{Few-shot learning}: Achieving sufficient forward transfer opens the door to few-shot learning of later concepts. 
The Bongard problem shown in Fig.~\ref{fig:bp57} is a great example of learning with only a few samples, utilizing the knowledge learned earlier. 
See Sec.~\ref{sec:forwardtransfer} and \ref{sec:rd_bongard} for details.

\textbf{Backward transfer}: 
This is knowledge transfer from $T_{\geq i}$ to $T_{<i}$ -- the opposite direction as forward transfer.  
When learning new tasks $T_{j}$ they may in turn help to improve the performance of $T_{i <j}$. 
This is like an ``overall review'' before a final exam after materials of all chapters have been taught and learned. Later materials can often help better understand earlier materials. See Sec.~\ref{sec:backwardtransfer} for details.

\textbf{Adapting to concept update/drift}: 
This is the ability to continually perform well at a given task $T_{i}$ by utilizing additional training data for $T_{i}$ if it arrives at a later time (concept updating), or when new data for the changing environment arrives (concept drift).  This process should only use the data of $T_{i}$ to update the model while not forgetting other learned tasks. See Sec.~\ref{sec:conceptdrift} for details.


\textbf{Non-confusion}: 
Machine learning algorithms often find the minimal set of discriminating features necessary for classification.  Thus, when more tasks emerge for learning in our LLL setting, earlier learned features may not be sufficient, leading to confusion between classes.  
For example, after learning to distinguish between images of ``1'' and ``0'' as $T_{1}$ and $T_{2}$, the learned model may identify  straight stroke for class ``1'' and curved stroke for ``0''.  But after learning ``I'' and ``O'' as $T_{10}$ and $T_{11}$ for example without using data for ``0'' and ``1'', the model may confuse between $T_{1}$ and $T_{10}$ (``1'' and ``I'') and similarly between $T_{2}$ and $T_{11}$ (``0'' and ``O'') when the model is tested on all tasks learned so far.
In human lifelong learning, this type of confusion may happen too.
For example, when meeting new people, if the first two people are visually distinct (such as very tall vs. very short) we can rely on this feature to tell them apart. However, if more people arrive and they are similar to the first two people, we may initially confuse them and must find finer details to reduce confusion, or to uniquely
distinguish them. In the extreme case where we encounter identical twins, significant effort may be required to learn the necessary details (by re-using their facial image data). See
Sec.~\ref{sec:confusionreduction} for details.


\textbf{Human-like learning}: As a new type of evaluation criteria for LLL, we can consider how well an approach is predictive of human learning behaviour.
If a LLL approach or framework is also able to provide explanatory power and match peculiarities of human learning (such as confusion or knowledge transfer in certain scenarios), it would have value in fields outside of machine learning. We discuss such connections between our framework and human learning in Section~\ref{sec:parallels}.

\subsection{Comparison of Different LLL Approaches}
\label{sec:comparison}

The mechanisms used to perform LLL tend to fall into three categories and often 
only demonstrate subsets of LLL properties as previously discussed. 
The first mechanism, replay, commonly works by storing previous task data and training on it alongside new task data \cite{rebuffi2017icarl, isele2018selective, chaudhry2019continual, wu2019large}. As a result of its data and computation inefficiency, we consider it generally not to be very a human-like learning mechanism. In our framework, we discuss particular cases where it is useful to combine replay with our central mechanism (Sec.~\ref{sec:backwardtransfer}, \ref{sec:confusionreduction}, \ref{sec:gracefulforgetting})

The second mechanism is regularization. This mechanism works by restricting weight changes (making them less ``flexible'') via a loss function so that learning new tasks does not significantly affect previous task performance \cite{kirkpatrick2016EWC, zenke2017synaptic,chaudhry2018riemann,ritter2018online,li2017learning, zhang2020class}. We use this mechanism as the basis for our unified framework. However, instead of simply controlling weight flexibility to \textit{retain} previous task performance, we leverage it to also encourage forward and backward transfer (Sec.~\ref{sec:forwardtransfer} and \ref{sec:backwardtransfer}), adapt to concept drift (Sec.~\ref{sec:conceptdrift}), reduce confusion (Sec.\ref{sec:confusionreduction}), and perform graceful forgetting (Sec.~\ref{sec:gracefulforgetting}).

The third mechanism, dynamic architecture, commonly works by adding new weights for each task and only allowing those to be tuned \cite{rusu2016progressive, yoon2018lifelong, xu2018reinforced}. This is often done without requiring previous task data and completely reduces forgetting while also allowing previous task knowledge to speed up learning of the new task. While this mechanism is necessary for LLL of an arbitrarily long sequence of tasks (any fixed-size network will eventually reach maximum capacity), it should be used sparingly to avoid unnecessary computational costs. In Sections~\ref{sec:expansion}, \ref{sec:forwardtransfer}, \ref{sec:conceptdrift}, and \ref{sec:confusionreduction}, we describe how dynamic architectures can be efficiently utilized to help achieve many LLL properties by combining it with our central mechanism.

Table~\ref{tab:skills} compares previous LLL approaches based on the three mechanisms with our unified LLL framework.

\input{tables/skills}

%% file: tables/skills.tex
\begin{table}[h]
    \centering
    \resizebox{\textwidth}{!}{%
        \begin{tabular}{llllll}
        \hline
        \textbf{Property} & \textbf{Replay-based} & \textbf{Regularization-based} & \textbf{Dynamic architecture-based} & \textbf{Our framework} & \textbf{Human learning} \\ \hline
        \textbf{Continual learning and testing} & good & good & good & good & good \\
        \textbf{Non-forgetting} & good & poor/okay & good & good & good \\
        \textbf{Forward transfer} & okay/good & okay/good & good & good & good \\
        \textbf{Few-shot learning} & poor & poor/okay & poor & good & good \\
        \textbf{Backward transfer} & good & poor & poor/okay & good & good  \\
        \textbf{Adapting to concept update/drift} & good & poor & poor/okay & good & good  \\
        \textbf{Non-confusion} & good & poor & poor & good & good  \\
        \textbf{Human-like learning} & poor & poor/okay & poor & good & good  \\ \hline
        \end{tabular}%
    }
    \caption{A rough assessment of the skills exhibited by the different main types of LLL mechanisms (replay, regularization, and dynamic architectures), as well as our new conceptual framework and human learning. Where multiple value are given, the performance is mostly approach-specific. For our framework, ``good'' means expected to be good, or conceptually designed to be good. Future research is required to demonstrate true performance. With regard to the performance of human learning, in Sec.~\ref{sec:parallels} we discuss how not all people may display these properties all the time.}

    \label{tab:skills}
\end{table}

%% file: sec_3_framework.tex
\section{A Unified Framework for Lifelong Learning }
\label{sec:framework}

In this section we describe how our unified framework works. We start by introducing the central mechanism and in the rest of the section, discuss how to use the central mechanism and combine it with additional mechanisms to achieve the several desirable LLL properties described in Sec.~\ref{sec:properties}. In the pseudo-codes provided, the lines where the central mechanism is applied is marked by a \textcolor{Cerulean}{$\CIRCLE$}.

While not restricted to a particular neural network type, we primarily consider our framework as applied to deep neural networks, which have become popular in recent years, and are an attractive type of machine learning model due to their ability to automatically learn abstract features from data.

\subsection{A Central Consolidation Mechanism}

We propose a lifelong learning framework which situates a consolidation policy as the central mechanism. The consolidation policy works through a single dynamic hyperparameter, $\pmb{b}$, 
which separately controls the flexibility of \textit{all} network weights. 
In other words, each weight in the network can have its own consolidation value, representing how easy (or hard) it is to modify the weight. 
Depending on the specific $\pmb{b}$-setting policy used during training, several desirable learning properties can be achieved. While the individual weights of the network are learned via back-propagation, $\pmb{b}$ is set by a consolidation policy.

The central consolidation mechanism ultimately works through dynamically modifying the loss function. To describe more precisely, we will deviate a little from the notation used in \cite{kirkpatrick2016EWC}, which introduced the Elastic Weight Consolidation LLL approach. If each network weight, $\theta_{i}$, is associated with a consolidation value of $\pmb{b}_{i} \geq 0$, the loss for the new task by itself, $L_{t}$, is combined with weight consolidation as follows:

\begin{equation}
    L(\theta) = L_{t}(\theta) + \sum_{i}\pmb{b}_{i}(\theta_{i}^{t} - \theta_{i}^{target})^{2}
\label{eqn:loss}
\end{equation}

Here, $\theta_{i}^{target}$ is the target value for a weight to be changed to. This can be either its value before training of the new task, or zero, in the case where we explicitly want to prevent certain weights from being used. $\theta_{i}^{t}$ is the weight value being updated during training on task $t$. 
The loss now had the following behaviour: a large $\pmb{b}_{i}$ value indicates that changing weight $\theta_{i}$ away from $\theta_{i}^{target}$ is strongly penalized during training. When a set of weights have large corresponding $\pmb{b}$ values, we will often refer to them as \textbf{``frozen''}. In contrast, value of $\pmb{b}_{i} = 0$ indicates that the weight is free to change. We refer to these weights as \textbf{``unfrozen''}. If $\pmb{b}_{i}$ is arbitrarily large, we can consider $\theta_{i}$ to be masked during backpropagation and completely prevented from changing to improve efficiency.



\subsection{Continual Learning of New Classification Tasks}
\label{sec:expansion}

In both lifelong and human learning, we desire to learn new tasks after learning previous tasks. In humans, this is enabled by the ability to continually grow new connections between neurons and remove old connections. When this ability is compromised, so is our ability to learn new things. Similarly, in our conceptual framework we consider learning new tasks with the help of network expansion and pruning.

The pseudo-code in Algorithm~\ref{alg:new_task} describes how to learn a new task, $T_{k}$, in a deep neural network in our conceptual framework after previous tasks $T_{1},...T_{k-1}$ have been learned. The role of the consolidation policy in this algorithm (and all others in this paper), as marked by \textcolor{Cerulean}{$\CIRCLE$}, is to ensure that newly added weights have the proper flexibility and that previous task weights will be inflexible, to prevent forgetting (Sec.~\ref{sec:nonforgetting}).

\begin{algorithm}[h]
\SetAlgoLined
    \tcp{Given that tasks $T_{1}, ..., T_{k-1}$ have been learned}
    Recruit free units for $T_{k}$ \tcp{this can be all available units, or determined by the task difficulty and similarity with previous tasks. See Sec.~\ref{sec:forwardtransfer}}
    \textcolor{Cerulean}{$\CIRCLE$} Initialize weights from earlier units to newly recruited units \tcp{see green links in Fig.~\ref{fig:framework_p1} and relating to forward transfer, see Sec.~\ref{sec:forwardtransfer}}
    \textcolor{Cerulean}{$\CIRCLE$} Initialize weights for the new units \tcp{see blue links in Fig.~\ref{fig:framework_p1}}
    \textcolor{Cerulean}{$\CIRCLE$} Set consolidation values for non-forgetting of previous tasks \tcp{see red links in Fig.~\ref{fig:framework_p1}. For non-forgetting, see Sec.~\ref{sec:nonforgetting}}
    Train the new task $T_{k}$ to minimize Eq.~\ref{eqn:loss}  \tcp{only on the data of new task $T_{k}$}
    \textcolor{Cerulean}{$\CIRCLE$} Optionally prune the network for $T_{k}$ \tcp{prune to free units for future tasks and graceful forgetting (Sec.~\ref{sec:gracefulforgetting}). }
\caption{Continual Learning of New Tasks (\textcolor{Cerulean}{$\CIRCLE$} indicates consolidation policy)}
\label{alg:new_task}
\end{algorithm}

This is a very general, high-level framework, and in subsequent sections, many of the steps in Algorithm~\ref{alg:new_task} will be explained and expanded upon.  On the topic of network expansion and pruning, a lot of previous work has been done and more future research is possible:

\begin{figure}[ht]
    \centering
    \includegraphics[width=0.9\textwidth]{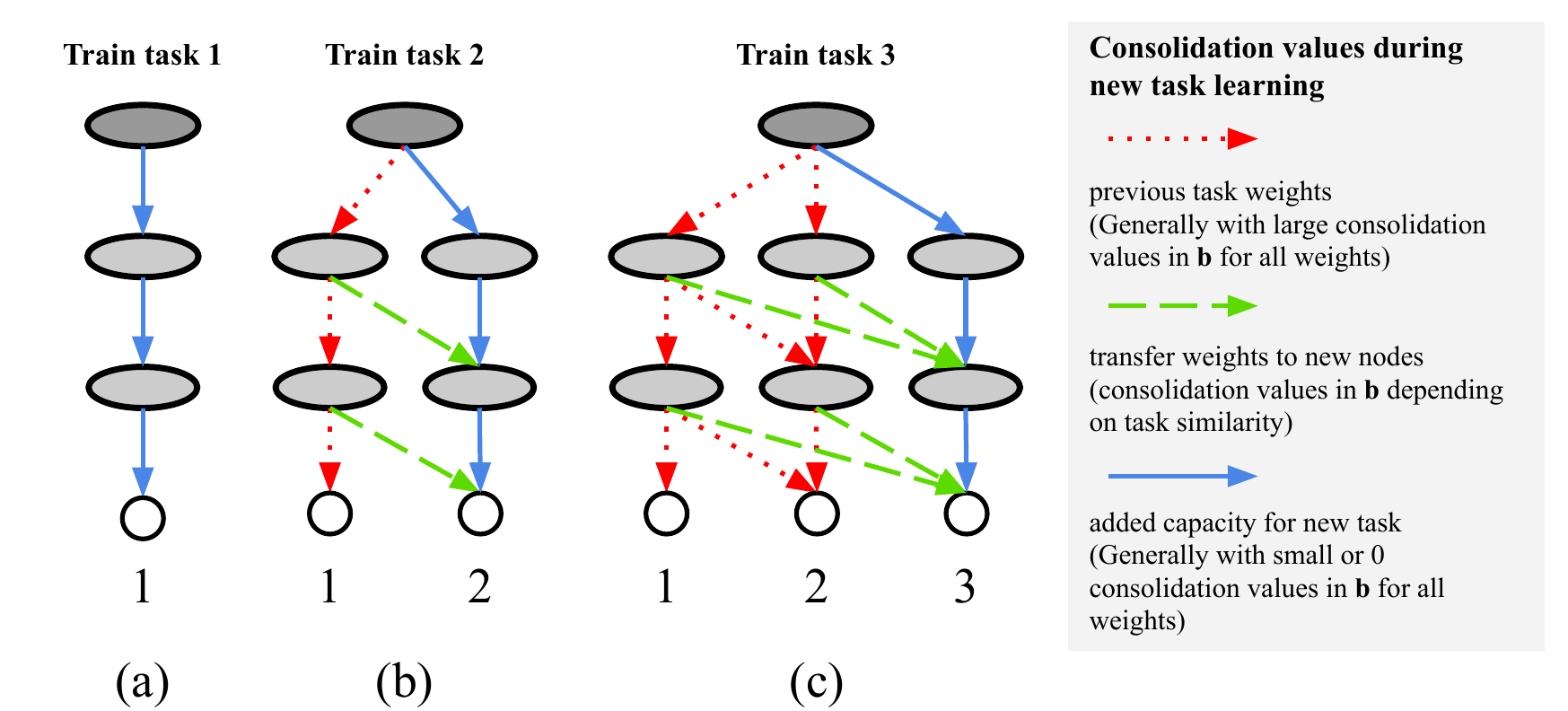}
    \caption{Network expansion and consolidation in our LLL framework. In step (a), task 1 is being trained. In step (b) task 2 is being learned without forgetting of task 1. In (c), task 3 is learned without forgetting of the previous tasks. Note that links represent the weights between groups of nodes and missing links indicate disconnected nodes (weights frozen with value of zero).}
    \label{fig:framework_p1}
\end{figure}

\begin{figure}[ht]
    \centering
    \includegraphics[width=0.75\textwidth]{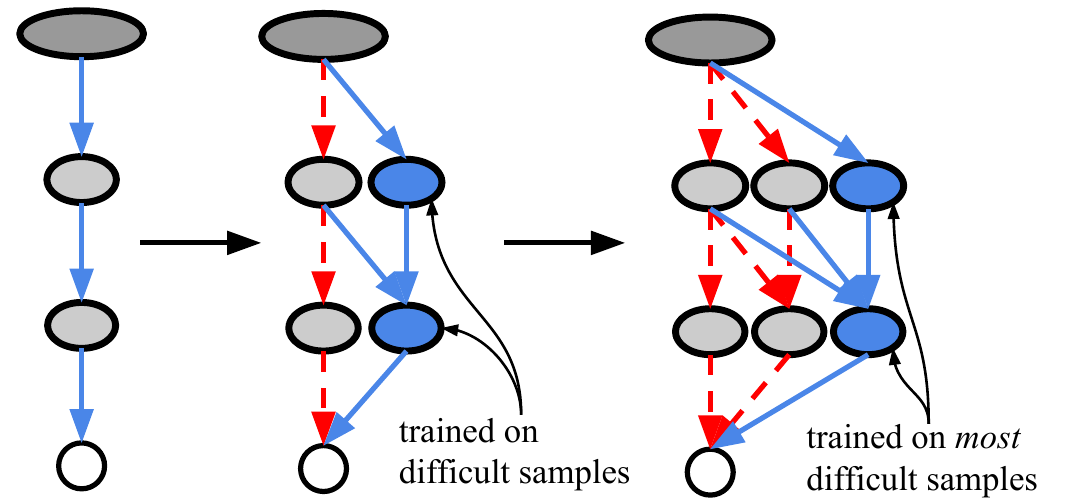}
    \caption{A visualization of multi-stage training of a single task. Samples that have thus far not been learned are used to train a constant number of new nodes, repeated until some criteria is met. Dashed red links are frozen while blue are free to be tuned}
    \label{fig:boosting}
\end{figure}

\begin{enumerate}

    \item Determining precisely how and by how much to expand a network is a difficult problem. For network expansion, several potential solutions which strike various balances between computational and memory efficiency and performance are as follows:
    
    \begin{enumerate}
        \item Do not add any new nodes and always tune the whole network on each new task \cite{kading2016fine}.
        \item Add a constant number of new nodes per task \cite{rusu2016progressive}.
        \item Estimate the required additional size \cite{ostapenko2019learning}. Complex tasks (according to some metric) will likely require greater capacity. If we identify the new task as being similar to previous ones, it may require fewer additional nodes to be added, helping avoid an expensive expansion, compression, or fine-tuning process \cite{yoon2018lifelong, li2019learn}. See Section~\ref{sec:forwardtransfer} for more details.
        \item Extend the network with an arbitrary computational graph. While most works consider extending individual layers of the existing architecture \cite{rusu2016progressive, yoon2018lifelong, li2019learn}, this is not strictly necessary. While easier tasks may be solved with fewer layers, more difficult ones may require additional depth, which is not considered by previous dynamic architecture approaches. See Sec.~\ref{sec:forwardtransfer} for how to potentially apply this idea to solving Bongard problems for example.
    \end{enumerate}
    
    \item In the case of dynamic architectures, the expanded model size also increases computation costs. To accommodate new task learning, pruning of weights or nodes used by previous tasks may also be performed \cite{golkar2019continual}. Related research directions and interesting ideas related to this are as follows:
    
    \begin{enumerate}
        \item Recent interesting work on the ``lottery ticket hypothesis'' \cite{frankle2018lottery} demonstrates that sparse sub-networks can be found which retains (and sometimes even improves) task performance after a high degree of pruning.
        
        \item In our lifelong learning framework, network pruning (at the node instead of weight level) also means more free units can be allocated for the future tasks. The authors of \cite{golkar2019continual} emphasize the importance of differentiating between weight and node pruning, and show that by pruning at the node instead of weight level \cite{yoon2018lifelong}, non-forgetting and efficiency can be further improved.
        
    
    \end{enumerate}
    
    \item Multiple steps of network expansion may be used to learn a task in stages. In each stage, a fixed network capacity may be recruited and trained. The still incorrectly classified samples may be trained with the expanded network at the next stage.  This idea is shown in Figure~\ref{fig:boosting}. The segmentation of samples learned in stages may correspond to different difficulty levels of the data. 
    In \cite{zhang2016understanding} the authors demonstrated that deep neural network can perfectly fit to random labels or random pixels given enough training time, but tend to fit to less random data earlier. In our framework, the easiest and ``clean'' samples would be learned at earlier stages, and more difficult or noisy samples would be learned in later training stages.  It would be very interesting future research to show this effect in our framework. 

    \item It would additionally be interesting to understand how our dynamic expanding models with various consolidation policies (to be discussed in this section) can be implemented efficiently. As training is applied on a set of one task at a time, and forward- and backward transfer may help reduce training, it would be interesting to see if this LLL framework can actually be computationally effective. Note that this would not be simply measured by the GPU time when the LLL framework is implemented.  We need to measure the ``energy'' used, to be more comparable to the amount of human brain power needed in similar lifelong learning. See Sec.~\ref{sec:parallels} for discussion on similarities between our framework and human learning.

\end{enumerate}

\subsection{Non-Forgetting}
\label{sec:nonforgetting}

Maintaining performance on previous classification tasks while learning new concepts is the primary difficulty lifelong learning approaches aim to combat. In our framework, we can design consolidation policies to make sure previous tasks will not be forgotten while the new task is learned with the data for the new task only. 
An intuitive way to prevent forgetting is by using a larger $\pmb{b}$ value for weights which most influence the loss of a trained model \cite{kirkpatrick2016EWC}. 

Our framework is inspired by the influential Elastic Weight Consolidation (EWC) \cite{kirkpatrick2016EWC}, which essentially 
uses the second derivative of the loss with respect to each weight as the weight consolidation value. 
Intuitively, well-learned previous classification tasks usually have a sharp local minimum (compared with a randomly-initialized network for the new task), thus, have relatively large second derivatives and $\pmb{b}$ values. 
However, in our framework, $\pmb{b}$ can be set to any large values based on the policy (to be designed), which controls how much non-forgetting should occur, based on desired applications. 
If $\pmb{b}$ is set to infinitely large, then forgetting would not happen at all, and this is equivalent to training a separate new network for the new classification task. 


Pseudo-code reflecting how a consolidation policy can simply be used in our framework is given in Algorithm~\ref{alg:non-forgetting}. The role of the consolidation policy here, similar to in Algorithm~\ref{alg:new_task}, is to prevent weights for previous tasks from being changed, but allowing the new task weights to change.

\begin{algorithm}[h]
\SetAlgoLined
    \tcp{We need to learn a new task $T_{k}$ by using data for only $T_{k}$, without forgetting $T_{1}...T_{k-1}$}
    \textcolor{Cerulean}{$\CIRCLE$} Unfreeze only units for $T_{k}$ \tcp{blue and green links in Fig.~\ref{fig:framework_p1} with $\pmb{b}$ set to small values}
    \textcolor{Cerulean}{$\CIRCLE$} Freeze all weights for tasks $\{T_{1}...T_{k-1}\}$ \tcp{red links in Fig.~\ref{fig:framework_p1}, with $\pmb{b}$ set to large values}
    Train the new task $T_{k}$ to minimize Eq. \ref{eqn:loss} 
\caption{Non-forgetting}
\label{alg:non-forgetting}
\end{algorithm}

The possible space of such consolidation policies is wide-ranging. In reference to existing work, examples of consolidation policies for non-forgetting with varying levels of effectiveness are given below, along with possible research directions. 

\begin{enumerate}
    \item When training only new tasks with regular SGD, no consolidation is used, reflecting $\pmb{b} = \pmb{0}$. But this causes previous tasks be forgotten, requiring retraining to maintain performance. This is exactly what LLL strives to prevent.
    
    \item In EWC \cite{kirkpatrick2016EWC}, the consolidation policy computes $\pmb{b}$ with the Fisher information matrix, as previously mentioned.
    
    \item In Progressive Networks \cite{rusu2016progressive}, where the network is extended for each task and previous task weights are masked during training, we can consider the policy to place infinite consolidation weight on the old part of the network, and zero on all new weights.
    
    \item The consolidation policy can be \textit{learned}, instead of manually engineered as in the previous examples. That is, instead of defining a fixed algorithm which computes the importance of weights, we can learn how to compute importance. This idea is considered in meta-continual learning \cite{vuorio2018meta}. In this work, one network learns to predict the consolidation policy of another network. 
    
    \item In data-focused regularization methods, a distillation loss is often employed. This loss ensures that outputs corresponding to previous tasks do not change (not necessarily evaluated with corresponding previous tasks' data) \cite{li2017learning, zhang2020class}.
    
    \item It would be an interesting future research to derive a theoretical bound on the forgetting effect on the previous tasks given $\pmb{b}$.

\end{enumerate}


\subsection{Forward Transfer}
\label{sec:forwardtransfer}


While Section~\ref{sec:nonforgetting} ensures previous tasks will not be forgotten during learning the new task, the previous tasks do not ``help'' learning the new task, a concept prominent in multi-task and transfer learning \cite{pan2009survey, zhang2017survey}, and appears in LLL as ``forward transfer''.
However, the question of how to utilize the previous task skills to accelerate new learning is a difficult and ongoing research question. In this conceptual framework discussion, 
we refer to the function estimating how much one task can be expected to help learn another as simply ``similarity'', $sim: \mathcal{T}\times \mathcal{T} \longrightarrow [-1, 1]$. This function is not necessarily commutative (i.e. $sim(T_{i}, T_{j}) \not\equiv sim(T_{j}, T_{i})$\footnote{For example, consider the case where one task is actually a subset of another. Here, the more general task will be more helpful for the smaller task.}).  If a new task is very similar (or identical in the extreme case) according to some metric such as visual similarity, ($sim \approx 1$) then the new task can be learned with no or little training data (few-shot learning). In other cases, previous tasks may actually impair the new task learning ($sim < 0$).

As summarized in Algorithm \ref{alg:forward_transfer}, positive forward transfer in our framework may be achieved through initialization of the weights for current task based on its similarity with previous tasks. 
For example, when initializing the output-layer weights for a new task, if the new task $T_{k}$ is identical to a previous task $T_{j}$, then we can initialize the weights of $T_{k}$ to reflect the values of those for $T_{j}$, requiring no or few new units for $T_{k}$. This idea is conceptually similar to that used by GO-MTL \cite{kumar2012learning} and ELLA \cite{ruvolo2013ella}, where knowledge is selectively shared between related tasks. The role of the consolidation policy, discussed later in this subsection, is to further allow positive transference and limit negative transference by controlling flexibility of weights between tasks (green links in Fig.~\ref{fig:framework_p1}).

\begin{algorithm}[h]
\SetAlgoLined
    \tcp{Assume $T_{1}, ..., T_{k-1}$ have been learned, and we need to “transfer” their knowledge while learning the new task $T_{k}$ }
    $\mathbf{s}$ = $(sim(T_{1}, T_{k}), ..., sim(T_{k-1}, T_{k}))$\;
    \textcolor{Cerulean}{$\CIRCLE$} Initialize weights for $T_{k}$ based on $\mathbf{s}$ \tcp{as weight initialization, see Fig.~\ref{fig:fwd_transfer}}
    \textcolor{Cerulean}{$\CIRCLE$} Set values of $\pmb{b}$ for $T_{k}$’s links based on $\mathbf{s}$\;
    Train the new task to minimize Eq.~\ref{eqn:loss} 
\caption{Forward Transfer}
\label{alg:forward_transfer}
\end{algorithm}

\begin{figure}[ht]
    \centering
    \includegraphics[width=0.7\textwidth]{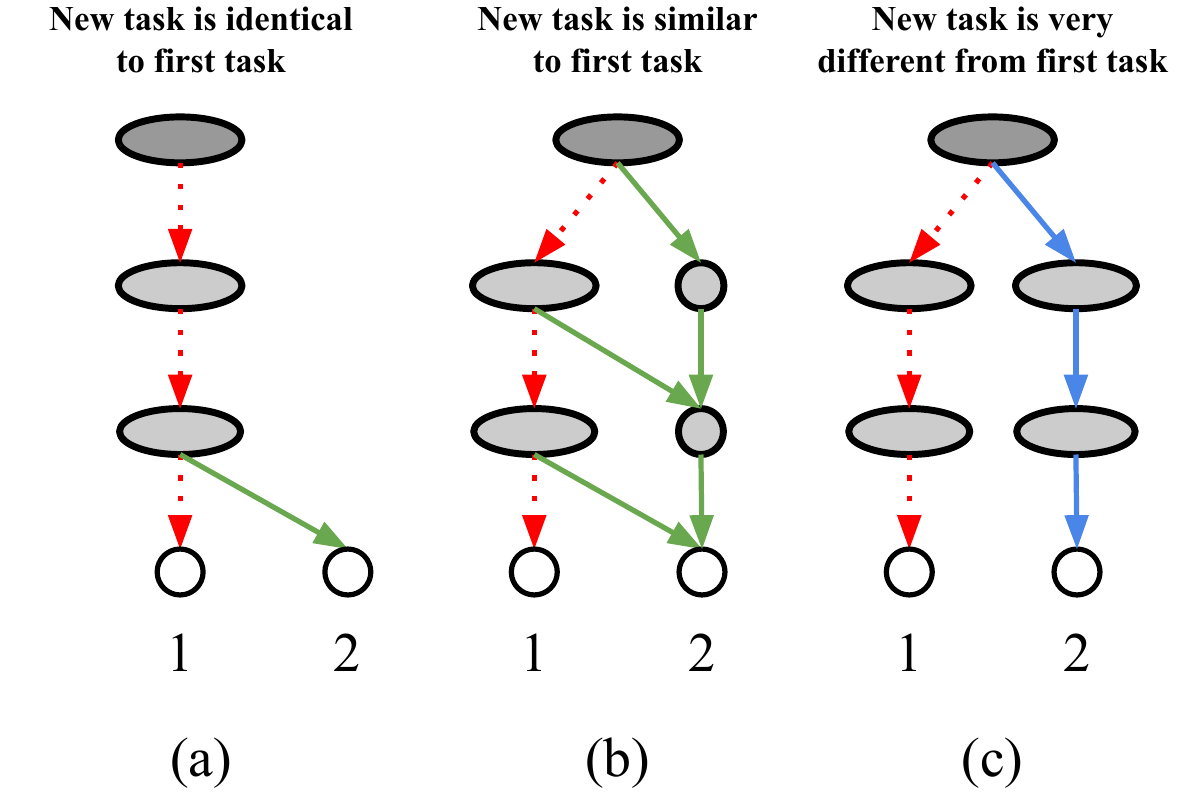}
    \caption{Weight initialization for different forward transfer cases. Green weights indicate the the weights are initialized to reflect those of the red weights with large values in $\pmb{b}$. Blue weights are randomly initialized and have small values in $\pmb{b}$. The width of the hidden layers reflect the relative number of nodes. In (a) is a special case where $T_{2}$ is identical to $T_{1}$ so no free units are needed, and weights can be ``copied''. In (b) is a case where $T_{2}$ is similar to $T_{1}$ in terms of training data or domain knowledge needed, so that a smaller number of free nodes are needed and weights of $T_{2}$ could be partially copied from $T_{1}$.  In (c) is when negative transfer from $T_{1}$ to $T_{2}$ may happen or when tasks are unrelated.}
    \label{fig:fwd_transfer}
\end{figure}

Ideas for research with respect to forward transfer in our framework include the following:
\begin{enumerate}
    \item After initializing the new weights for the new task, we can consider taking a further step in avoiding negative forward transference with proper consolidation (line 3 of Algorithm \ref{alg:forward_transfer}). Thus identifying when and how negative transfer would occur from a previous task to a new one is an interesting existing research problem \cite{wang2019characterizing}.

    \item How to initialize the new weights for a task given similarity to other tasks is an important problem related to meta-learning. As already mentioned, if a task is identical to a previous one, weights may be copied, however, if the new task is otherwise ``correlated'' with a previous task (for example, task 1 predicts how much an animal weights and task 2 predicts how portable animals are as pets), there may also be a way to efficiently utilize the previous task weights. This would allow minimizing network size for a given desired performance level by avoiding parameter redundancy in an efficient manner (in contrast to the prune-and-retrain technique \cite{yoon2018lifelong, li2019learn}).
    
    \item When a new task is sufficiently similar to a previously learned task, it is natural to consider how few examples are required to learn the new task well, taking us into the regime of few-shot learning \cite{fei2006one}, an extremely important research direction which we also touch on in Section~\ref{sec:rd_curriculum}.  
    
    
    \item Lottery tickets (LTs) have been show to generalize well to new tasks \cite{morcos2019one}. This suggests that one potential way of achieving forward transfer in a meta-learning fashion via weight initialization is with reusing the LT identified for a previous task. Although obtaining the LT for a task may be expensive, the time saved in training later tasks may result in an overall benefit. This idea of meta-learning by copying useful neural structures and weight initialization to accelerate adapting to new functions is also hypothesized to occur during evolution of biological brains \cite{chakraborty2015brain}. This would be a fascinating future research topic. 
    
    \item This conceptual framework may provide an additional perspective on multi-task reinforcement learning (RL) \cite{oh2017zero, DEramo2020Sharing, rusu2016progressive}. While RL in a LLL setting with dynamic architectures has been considered before \cite{rusu2016progressive}, our framework adds the additional concept of maximizing positive transfer and avoiding negative transfer between tasks \cite{DEramo2020Sharing}, increasing the chances of successful few-shot learning.
    
    \item A perspective on Federated Learning \cite{smith2017federated} is also provided by this framework when multiple new tasks are trained remotely and later their weights are forward-transferred to new tasks. 
    For example, when a new device begins training on a task with its private data, the weights of previously trained devices can be ``forward-transferred'', without revealing the private data. 
    Future research can be conducted to see how our framework be utilized in federated learning.  
    

\end{enumerate}

\subsection{Backward Transfer and Overall Refinement}
\label{sec:backwardtransfer}

\begin{figure}[ht]
    \centering
    \includegraphics[width=0.3\textwidth]{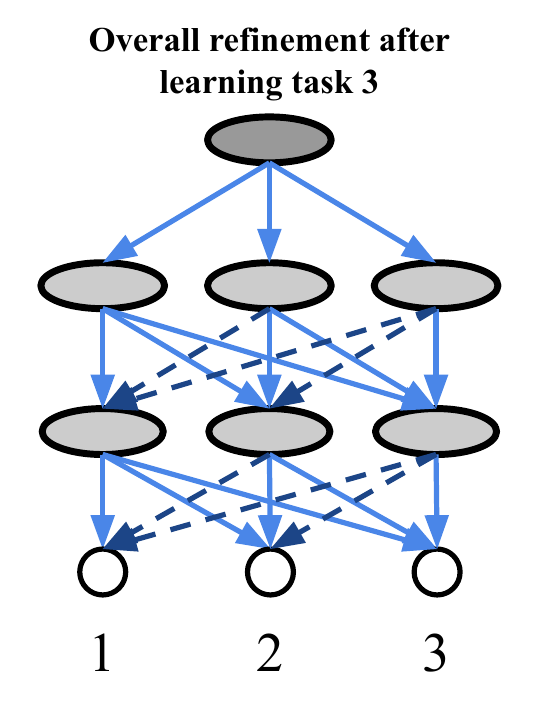}
    \caption{Consolidation of weights during overall refinement. All weights now have $\pmb{b} = 0$ and are able to be tuned together, including the random-initialized backward transfer weights (dashed dark blue links). This process is essentially the same as normal batch learning where data from all tasks is used, but with continual learning to ``set'' the initial weights. Since forward transfer has already been taken advantage of (Alg.~\ref{alg:forward_transfer}) and each task has been trained separately, this process would require little time.}
    \label{fig:overall_refinement}
\end{figure}

In Section~\ref{sec:forwardtransfer} we discussed forward transfer where previous tasks help to learn the new task. Can the new tasks similarly help to improve performance on previous tasks to achieve positive backward transfer?
If samples are available for previous tasks, our framework may be able to achieve such backward transfer as well as overall refinement (including forward transfer and resolving remaining confusion) by unfreezing the entire network before tuning on all tasks (as in Figure~\ref{fig:overall_refinement}) with stored samples (also called ``rehearsal''). The corresponding pseudo-code for this concept is in Algorithm~\ref{alg:overall_refinement}. The consolidation policy in this algorithm simply unfreezes the entire network to allow for tuning, however, more complex policies could be used (as mentioned later in this subsection and earlier in Section~\ref{sec:forwardtransfer}).
\begin{algorithm}[h]
    \SetAlgoLined
    \tcp{Some data for tasks $T_{1}, ... T_{k}$ is needed here}
    \textcolor{Cerulean}{$\CIRCLE$} Unfreeze the whole network with all connections ($\pmb{b} = 0$)\;
    Train all tasks to minimize Eq.~\ref{eqn:loss}\;
\caption{Overall Refinement (same as Batch Learning)}
\label{alg:overall_refinement}
\end{algorithm}

Rehearsal is one of the earliest strategies employed to combat catastrophic forgetting \cite{robins1995catastrophic}. In human learning, these approaches are analogous to periodically reviewing previously learned material or mixing in old facts with new ones. These methods store some subset of data for each task and use them during training of new tasks \cite{rebuffi2017icarl, isele2018selective, chaudhry2019continual, wu2019large}. Pseudo-rehearsal approaches, instead of storing samples, aim to maintain generator capable of producing novel samples from previous task distributions \cite{shin2017continual, atkinson2018pseudo}.

The process described in Algorithm~\ref{alg:overall_refinement} is similar to the more resource-intensive batch multi-task learning setting, where we assume that we can train on all tasks at once. However, in our LLL setting, if this process is only done infrequently after already taking care to achieve forward transfer and confusion reduction separately and more efficiently, the benefits of overall rehearsal can be obtained while minimizing the associated costs. 

Research directions to further develop these ideas include:

\begin{enumerate}
    \item It would be interesting to investigate whether the performance on earlier tasks ($T_{1},... T_{k-1}$), in terms of predictive accuracy, can actually be improved with backward transfer. This skill appears to be one of the more difficult ones to exhibit \cite{lopez2017gem, riemer2018learning}.
    \item It would be valuable to demonstrate, theoretically or empirically, that by learning with forward transfer, confusion reduction, and backward transfer, that we indeed reduce total computation and data costs compared to standard multi-task learning approaches.
    \item It would also be interesting to see if we can try take advantage of possible forward transfer and avoid negative transfer during overall refinement using ideas from Section~\ref{sec:forwardtransfer}, instead of completely unfreezing the entire network.
\end{enumerate}

\subsection{Adapting to Concept Update/Drift}
\label{sec:conceptdrift}

In the real world, we often do not learn a concept once and then receive no more useful information about it. For example, when we are young, we quickly learn to recognize houses. However, we continue to see new houses even after we have initially learned to recognize this. Humans are able to take these new observations and use them to slowly refine or \textit{update} our concept of what a house it. Similarly, the knowledge related to a single concept, such as a house, may change over time (``concept drift''). For example, when you are young, you learn to recognize the houses around you. However, upon travelling to another country or geographic region, houses may be made of different materials, be different sizes, or have other design differences. Humans can easily use these new experiences to update their understanding of what a house is. In Sec.~\ref{sec:rd_capsules} we discuss how the ``capsules'' for simpler concepts making up a house are updated after seeing this kind of new information.
Similarly, we can consider a LLL scenario where data for tasks may not arrive in such a clearly segmented order as described in Section~\ref{sec:setting} and where the input distribution may also change over time. Performing LLL where a possibly indefinite set of tasks is learned in a non-segmented way and in a non-stationary environment poses difficulties for existing LLL approaches however. Many approaches are unable to update previous task skills based on new data without causing forgetting in other tasks \cite{rusu2016progressive}. Existing research into this problem includes \cite{aljundi2019task}. 



In our conceptual framework, if new samples for an already-learned task arrive we can unfreeze the weights for $T_{k}$ and
resume training with the new data. If the existing architecture is not sufficient to adapt to the new data, free units can also be recruited (similar to Fig.~\ref{fig:boosting}). This idea is shown in Algorithm~\ref{alg:online_retraining}. The role of the consolidation policy in this algorithm is to ensure that only the weights for $T_{k}$ are modified during this training.

\begin{algorithm}[h]
    \SetAlgoLined
    \tcp{Assume new data is available for a previous task $T_{k}$}
    \textcolor{Cerulean}{$\CIRCLE$} Unfreeze units for $T_{k}$ \tcp{can gradually unfreeze a greater number of layers starting from output, based on degree of concept drift}
    \textcolor{Cerulean}{$\CIRCLE$} Freeze units for all other tasks\;
    Train task $T_{k}$ to minimize Eq.~\ref{eqn:loss} using new data for $T_{k}$\;
    \If{error on $T_{k}$ is not small enough}{
        Recruit free units for $T_{k}$ \tcp{similar to the process shown in Fig.~\ref{fig:boosting}}
        \textcolor{Cerulean}{$\CIRCLE$} Unfreeze only units for $T_{k}$ \tcp{see line 1}
        \textcolor{Cerulean}{$\CIRCLE$} Freeze units for all other tasks\;
        Train task $T_{k}$ to minimize Eq.~\ref{eqn:loss} using new data for $T_{k}$\;
    }
\caption{Adapting to Concept Update/Drift}
\label{alg:online_retraining}
\end{algorithm}

As a direction for future work, we can consider how the weights for a task should be updated as the amount of concept drift increases \cite{webb2016characterizing}. In the case of small changes, we can unfreeze only the output-layer weights for $T_{k}$, but where concept drift is larger, we can perform greater unfreezing of earlier layers as well (see Sec.~\ref{sec:rd_rand} for more on this idea). Additionally, perhaps in this case we can also re-evaluate task similarities to better make use of new transfer opportunities as tasks evolve.

\subsection{Non-Confusion}
\label{sec:confusionreduction}

\begin{figure}[ht]
    \centering
    \includegraphics[width=0.65\textwidth]{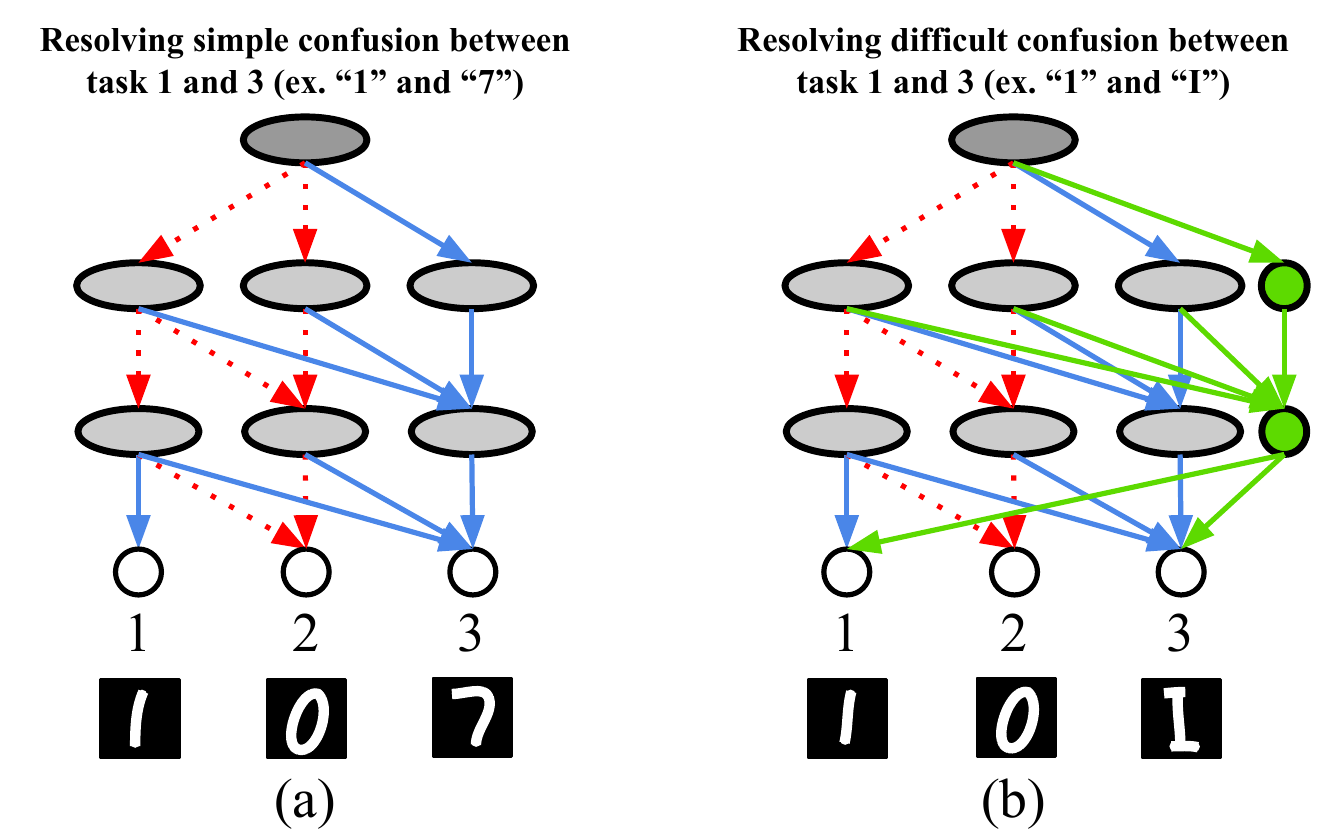}
    \caption{Network expansion during confusion reduction. Blue links are existing weights free to change during confusion reduction, and the green links and nodes are newly added to resolve confusion when necessary.}
    \label{fig:confusion_reduction}
\end{figure}

As discussed in Sec.~\ref{sec:properties}, if the sequence of tasks is learn to classify images of "1", "0", ..., "I", "O", confusion can happen between "1" and "I", and "0" and "O".  In fact, confusion can happen between any pair or subset of classification tasks\footnote{We are assuming that evaluation is done in a ``single-head'' manner where the task identity for a given test sample is unknown. This is in contrast to the easier ``multi-head'' evaluation \cite{chaudhry2018riemann}}.
Note that here we assume classes are mutually exclusive -- in the case of multi-label learning, certain ``confusions'' may be desired, and need not be resolved.
Here, we first fine-tune the network to reduce confusion. In cases where the current network is not sufficient to resolve confusion, network expansion can be considered, as demonstrated in Figure~\ref{fig:confusion_reduction}. 
In this algorithm, the role of the consolidation policy is to ensure that only those weights associated with the confused tasks are tuned, leaving other tasks unaffected.


\begin{algorithm}[h]
    \SetAlgoLined

    Use confusion matrix to select a subset of tasks, $\mathcal{T}_{confused}$, requiring confusion reduction \;
    \textcolor{Cerulean}{$\CIRCLE$} Unfreeze only units for $\mathcal{T}_{confused}$\;
    \textcolor{Cerulean}{$\CIRCLE$} Freeze units for all other tasks \tcp{see Fig.~\ref{fig:confusion_reduction}}
    Train $\mathcal{T}_{confused}$ to minimize Eq.~\ref{eqn:loss} using data of $\mathcal{T}_{confused}$\;
    \If{confusion not sufficiently resolved}{
        Recruit free units \tcp{see Sec.~\ref{sec:expansion} for expansion/pruning}
        \textcolor{Cerulean}{$\CIRCLE$} Unfreeze only units for $\mathcal{T}_{confused}$\;
        \textcolor{Cerulean}{$\CIRCLE$} Freeze units for all other tasks\;
        Train $\mathcal{T}_{confused}$ to minimize Eq.~\ref{eqn:loss} using data of $\mathcal{T}_{confused}$
    }

\caption{Confusion Reduction}
\label{alg:confusion_reduction}
\end{algorithm}

An interesting dilemma to note occurs when a new task $T_{k}$ is very similar to a previous task $T_{i}$. This is beneficial because forward transfer from $T_{i}$ to $T_{k}$ would greatly help learn $T_{i}$ quickly, as discussed in Section~\ref{sec:forwardtransfer}.  However, 
this would lead to large confusion between $T_{k}$ and $T_{i}$ which needs to be resolved later with samples for both $T_{k}$ and $T_{i}$. 
This situation is reminiscent of human learning where a similar kind of confusion can also happen (see the identical twin example in Sec.~\ref{sec:properties}) and can be resolved.



There are several possible lines of research in relation to confusion reduction in our framework:

\begin{enumerate}
    \item One interesting research problem is identifying the best way of obtaining or storing samples from previous tasks. These samples are useful when estimating the confusion matrix and when performing the fine-tuning. Several approaches for maintaining such a replay memory are discussed and compared in \cite{chaudhry2019continual}. Alternatively, a generative model, such as in \cite{shin2017continual, atkinson2018pseudo} can be used to generate new samples of previous tasks.
    
    \item How should we go about increasing free representational capacity for confusion reduction? Are there cases where we do need to extend the network, or is it sufficient to simply prune and re-assign weights, or perform graceful forgetting to free up capacity, as discussed in Section~\ref{sec:gracefulforgetting}.
    
    \item How should the consolidation policy behave during this stage? Instead of fully freezing or unfreezing weights, one option is to identify a smoother distribution over tasks to use for confusion reduction, in which case the consolidation of weights related to a task may be proportional to how badly it requires confusion reduction. 
    
    \item Can confusion be efficiently avoided during the initial training of a task by recognizing when it is likely to occur and mixing the corresponding previous task samples into the training data?
\end{enumerate}

\subsection{Graceful Forgetting}
\label{sec:gracefulforgetting}

As we learn more and more tasks and the network expands over time or the representational capacity is used up, we may reach a limit where either the network is unable to learn a new task without significantly modifying already-important parameters, or the maximum network size has been reached. In such cases, we can consider ``graceful'' forgetting \cite{aljundi2018memory} of previous tasks, $\mathcal{T}_{forget}$, to free up units or memory for learning new tasks. Three such methods of controlled graceful forgetting are as follows:

\begin{enumerate}
    \item Tasks of $\mathcal{T}_{forget}$ can simply be regarded as unimportant (if seldom used in deployment for example) and the corresponding values in $\pmb{b}$ decreased so that new tasks can make use of them without immediately harming performance of the to-be-forgotten task
    
    \item With graceful forgetting via pruning, as mentioned in Section~\ref{sec:expansion}, the amount of sparsification or forgetting done can be controlled with a simple threshold parameter for weight importance \cite{golkar2019continual}, which can be informed by acceptable performance loss or pruning amount. By lowering the acceptable performance on $\mathcal{T}_{forget}$ (e.g., lowering the predictive accuracy requirement from 95\% to 70\%), more nodes can be pruned and allocated to new tasks.
    
    \item We can reduce the amount of stored training samples, indirectly causing graceful forgetting through reduced rehearsal and freeing up memory for new tasks to use.
    
\end{enumerate}

\begin{figure}[ht]
    \centering
    \includegraphics[width=0.5\textwidth]{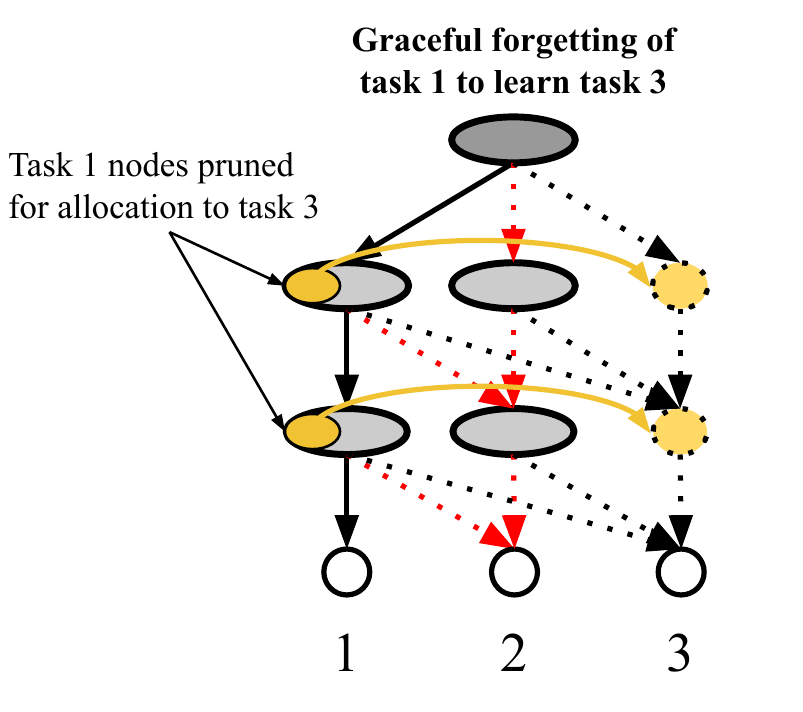}
    \caption{Graceful forgetting via node pruning of task 1 to accommodate task 3. Solid yellow links indicate the re-allocation of nodes previously used for task 1. In order to preserve task 2 performance, some link can still be frozen (dotted red links), but since task 2 may depend on task 1 nodes, some performance may be lost.}
    \label{fig:graceful_forgetting}
\end{figure}

In the pseudo-code of Algorithm~\ref{alg:slective_forgetting}, we consider the second option. The role of the consolidation policy in this algorithm is to ensure that while tuning on $T_{j}$ with a sparsity constraint, only the weights affecting $T_{j}$ are affected.

\begin{algorithm}[h]
    \SetAlgoLined
    \tcp{Assume task $T_{j}$ to forget}
    \textcolor{Cerulean}{$\CIRCLE$} Unfreeze units for $T_{j}$\;
    \textcolor{Cerulean}{$\CIRCLE$} Freeze units for all other tasks except $T_{j}$\;
    Train on task $T_{j}$ to minimize Eq.~\ref{eqn:loss} while under sparsity constraint\;
    Prune nodes used by $T_{j}$ to the desired threshold to free units \tcp{see Fig.~\ref{fig:graceful_forgetting}}

\caption{Graceful Forgetting}
\label{alg:slective_forgetting}
\end{algorithm}

As directions for future work, we can consider the following:
\begin{enumerate}
    \item Automatically deciding the rate at which forgetting of tasks should occur is an interesting problem. In a limited resource scenario, we may want to generally forget tasks when they are the least used, but also do not want to forget tasks which are important in supporting later tasks.
    \item How exactly should graceful forgetting be applied to convolutional filters or more complex architectures?
\end{enumerate}

%% file: sec_4_connections.tex
\section{Close Connections to Other Learning Settings}
\label{sec:connections}

In this section we will describe how our unified framework can be applied to learning settings beyond LLL and fully connected neural networks. Several of these learning settings are closely related to LLL in their setup and scope: multi-task learning, curriculum learning, few-shot learning, and transfer learning. We will provide a brief discussion of their connections to our framework. Other learning settings are based on the type of model being used: convolutional networks, deep random networks, and capsule networks. For each of these will will discuss how our framework could be applied to them or is otherwise connected to them.


\subsection{Multi-Task and Curriculum Learning}
\label{sec:rd_curriculum}

Special cases of LLL are multi-task learning \cite{caruana1997multitask} and curriculum learning \cite{bengio2009curriculum}. In multi-task learning, all $(T_{i}, D_{i})$ are provided together -- allowing the model to be trained on all tasks at the same time. In curriculum learning, all data is similarly made available, but the problem is to identify the optimal order in which to train on data for the most efficient and effective learning. 
An example of an intuitive type of curriculum is to learn tasks from ``easy'' to ``hard'' \cite{elman1993learning}, similar to the way humans often learn new concepts. This idea is reflected in Algorithm \ref{alg:curriculum_learning}.

\begin{algorithm}[H]
    \SetAlgoLined
    \tcp{Assume a pool of tasks to learn from}
    \While{pool is not empty}{
        Select the best task, $T_{best}$ \tcp{the easiest, for example}
        \textcolor{Cerulean}{$\CIRCLE$} Apply Alg.~\ref{alg:new_task} to learn $T_{best}$\;
        \textcolor{Cerulean}{$\CIRCLE$} Apply Alg.~\ref{alg:confusion_reduction} \tcp{reduce confusion if necessary}
        Remove $T_{best}$ from pool\;
    }
    \textcolor{Cerulean}{$\CIRCLE$} Apply Alg.~\ref{alg:overall_refinement} \tcp{final refinement of all tasks}
\caption{Curriculum Learning}
\label{alg:curriculum_learning}
\end{algorithm}

While research has been done on combining active and curriculum learning with LLL \cite{ruvolo2013active, sun2018active, de2019continual}, we suspect more is needed.
Research directions to further develop these ideas include:
\begin{enumerate}
    \item Understanding how a task selection mechanisms might work with a weight consolidation policy, where transference between tasks can be more explicitly controlled.
    \item Developing or learning to identify a taxonomy over samples to better construct curriculums such as simple to complex.
    \item Given that the general task type to be learned is known, can we automatically construct primitive pre-training samples. For example, if we are training a LLL more for hand-written characters, pre-training on simple strokes may help. 
    \item How much can an optimized curriculum reduce data requirements and improve few-shot learning performance?
\end{enumerate}


\subsection{Few-shot Learning and Bongard Problems}
\label{sec:rd_bongard}

\begin{figure}[ht]
    \centering
    \includegraphics[width=0.75\textwidth]{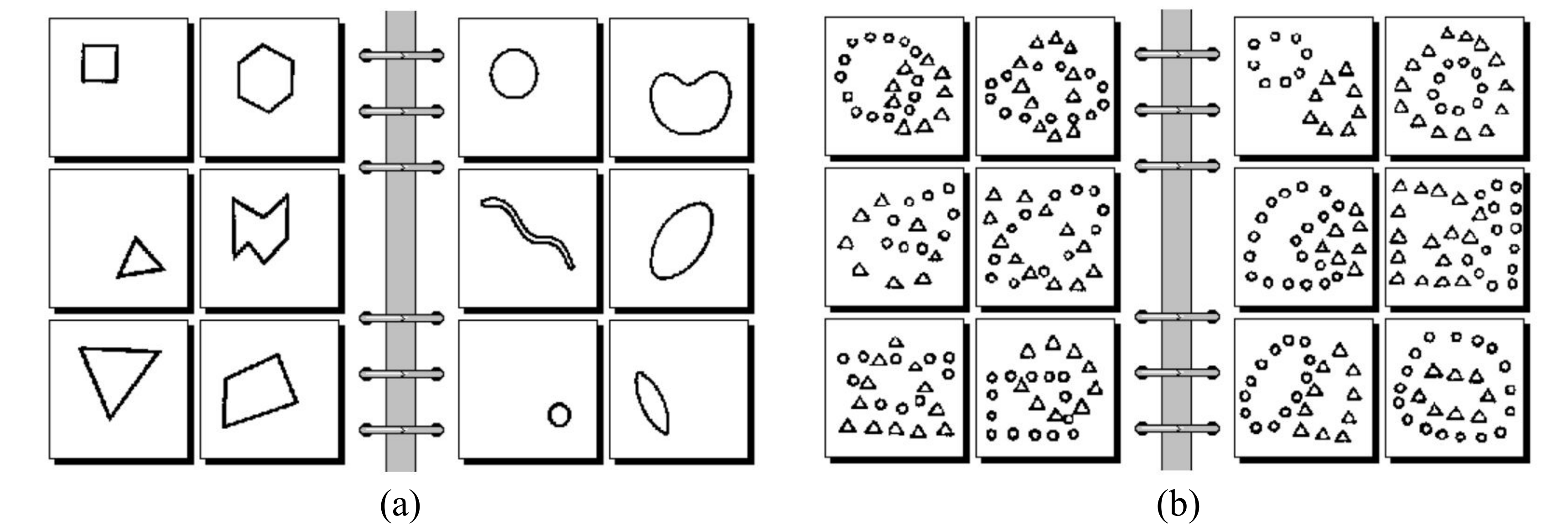}
    \caption{Two examples of Bongard problems. (a) BP \#5 is simple, where the rule is polygonal shapes on the left and curvilinear shapes on the right. (b) BP \#99 is quite hard, where the rule is that larger shapes formed by connecting similar small shapes overlap vs. the larger shapes do not overlap. Such highly abstract rules are common among BPs. For such difficult BPs (including that in Fig.~\ref{fig:bp57}), we find that for both humans and machine learning, adding more training samples often does not help, and rather it takes a ``stroke of insight'' or sufficiently similar previously acquired knowledge to solve it. We can easily generate more test data using the desired rule to test any learning algorithms after trained on a small number of data.}
    \label{fig:BP_examples}
\end{figure}

LLL and curriculum learning have great potential to achieve few-shot learning of difficult concepts, which humans are often capable of doing. By learning relevant easier concepts earlier with LLL, later harder ones could be learned with many fewer examples via forward transfer of knowledge. 

Humans can often discover the underlying rules and patterns of complex concepts with only a few examples.  
We can use Bongard Problems (BPs) \cite{bongard1967problem}
to illustrate this\footnote{They can be found here: \url{http://www.foundalis.com/res/bps/bpidx.htm}.}.
A particular BP asks a human to recognize the underlying rule of classification with  
only 6 examples on the left (say they are positive) and 6 examples on the right (negative samples).   
Two such BPs are shown in Figure~\ref{fig:BP_examples},
where (a) is a simple BP and (b) is a difficult BP for humans.
To solve BPs, humans often require identifying highly abstract rules learned earlier in their lives (or in easier BPs). 
Some form of LLL of many visual abstract concepts and shapes must likely have happened in humans so that they can accomplish BPs well with only 12 examples in total.  

\begin{figure}[ht]
    \centering
    \includegraphics[width=0.85\textwidth]{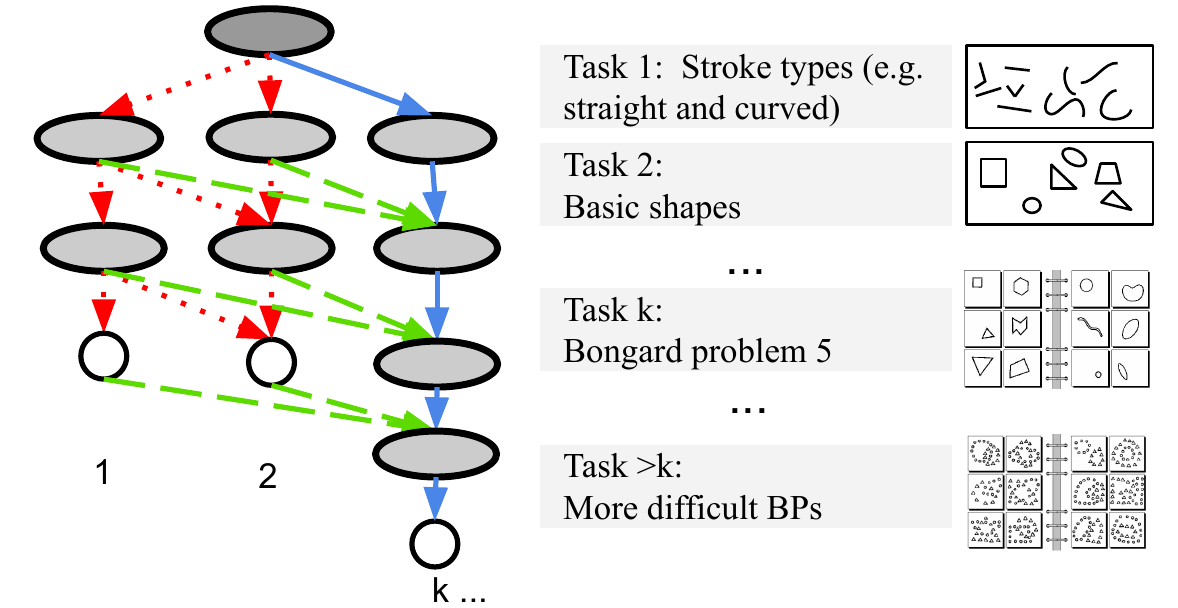}
    \caption{An illustrative example of curriculum learning to work up to solving Bongard problems with a small number of positive and negative examples. After learning to differentiate between basic strokes types, shapes, and possibly other early tasks, these representations including the output and hidden layers can be transferred to $T_{k}$ to possibly solving BPs such as BP\#5 (sharp angles vs. smooth curves) with a few examples. 
    This process can continue to allow us to solve a more complex visual reasoning problem with a few samples. Note that networks of subsequent tasks can have increasing depths in order to solve harder BPs after easier ones have been learned.}
    \label{fig:Curriculum_BP}
\end{figure}

We can outline a possible few-shot learning process to learn such BPs in our LLL framework, in a curriculum-like fashion. 
First, we can train (or pre-train) on simpler tasks to recognize simpler shapes.
Next, harder and more complex visual shapes and BPs would be 
trained. 
This idea of using curriculum learning to ``working up to'' more difficult BPs is reflected in Figure~\ref{fig:Curriculum_BP}. 
In this illustrative example, we feed previous class labels (the output nodes) into nodes of new tasks to provide opportunity for forward transfer of knowledge. With this approach, BPs could be solved with fewer examples, as shown in \cite{yun2020deeper}. 
Note that in this illustrative example, the networks of subsequent tasks can have different and increasing depths in order to solve more difficult problems after easier problems are learned.  

To solve difficult BPs with a few examples (such as BP \#99 in Fig.~\ref{fig:BP_examples}b), the importance of forward transfer of knowledge is crucial. If a person cannot solve BP \#99 (or other hard BPs) in several minutes, normally more training data would not help much.  Often ``a flash of inspiration'' or ``Aha!'' moment would suddenly occur and the problem is solved.  This is because a person has learned probably thousands or tens of thousands of concepts and abstract relations in his/her life, and the correct knowledge is suddenly combined to solve the hard BP.   

\subsection{Pre-Training Methods}
\label{sec:pretraining}

Pre-training is a ubiquitous process in deep learning \cite{devlinetal2019bert, girshick2014rich, wang2015unsupervised}. 
In computer vision (CV), nearly all state-of-the-art approaches rely on pre-training on a non-target task for which there is abundant data \cite{mahajan2018exploring}, such as the ImageNet dataset \cite{russakovsky2015imagenet}. Pre-trained CV models such as VGG-16 \cite{simonyan2014very} have been shared to the public allowing anyone to fine-tune their model to achieve otherwise infeasible performance on their target task. Due to the reduced need for target task training data, using pre-trained CV models is also a common approach to few-shot learning. \cite{ramalho2019empirical}.
In natural language processing, pre-trained language models have recently become popular \cite{conneau2017supervised, peters2018deep, devlinetal2019bert}. For example, BERT (Bidirectional Encoder Representations from Transformers) \cite{devlinetal2019bert}, has shown that state-of-the-art performance can be attained on several different NLP tasks with a single pre-trained model.

With pre-training, often a large neural network is first trained on one or more tasks with very large labeled training datasets.
The trained neural network would be then frozen, 
and a smaller network stacked on top of the large network and trained 
for improved performance on target tasks. From the perspective of our framework, the consolidation values of the smaller networks would be considered unfrozen for training.   

Pre-training can thus be viewed as certain weight consolidation policies in our LLL framework. 
The main difference is that most of our weight consolidation algorithms in the last section apply to weights per tasks, or column-wise in our figures, while for pre-training, consolidation is layer-wise. An interesting research topic would thus be to consider designing consolidation policies which perform both layer-wise and column-wise consolidation based on the task at hand.

A natural way to fine-tune the consolidation policy when tuning on the target task is to consider how much to unfreeze each layer. Instead of simply tuning the output layer or tuning the entire network, we can interpolate between these two policies, to allow 
gradual unfreezing of layers in the network to best adapt to the target task 
without losing the generalization benefits of pre-training.  See \cite{erhan2010does} for an example of such directions. 

Deep random networks are a another example of layer-wise consolidation policies.  See Sec.~\ref{sec:rd_rand} for more details. 


\subsection{Transfer Learning and Meta-Learning}
\label{sec:transferlearning}

Transfer learning is closely related to LLL as well as curriculum learning, few-shot learning, and pre-training. It is explicitly concerned with aiding the learning of a task by exploiting previously-acquired knowledge \cite{de2019continual}. With respect to LLL, it is solely concerned with forward transfer and not maintaining or improving previous task performance.

Aspects of our framework with application to pure transfer learning thus include continual learning (Sec.~\ref{sec:expansion}) and forward transfer (Sec.~\ref{sec:forwardtransfer}). After learning an initial set of tasks, being able to learn the new task may require expanding the network depending on the difficulty. The techniques discussed for forward transfer can be used to initialize these newly added weights as well as apply the appropriate consolidation to avoid negative transfer and maximize positive transfer.

Meta-learning is another learning setting similar to transfer learning. Instead of directly using previous task knowledge to help learn a known new task, we instead use the previous tasks to learn \textit{how} to learn \cite{hospedales2020meta} on an unknown new task. By optimizing the learning process, we can learn new tasks faster, with higher performance, and with less data. This kind of learning can occur in many ways, including:
\begin{enumerate}
    \item Learning a good parameter initialization. In Sec.~\ref{sec:forwardtransfer} we discuss methods for initializing new task weights to speed up learning.
    
    \item Identifying a good architecture. In Sec.~\ref{sec:expansion} we consider estimating the additional network size required to learn a new task to a sufficient performance level. This estimate can be based on both how difficult the task is as well as how similar it is to previous tasks.
    
    \item Learning to compute task similarity. If we perform meta-learning in preparation of learning \textit{many} tasks (as in LLL, multi-task learning, or continual learning), we can consider meta-learning the task similarity metric, $sim$ (Sec.~\ref{sec:forwardtransfer}). This would have many downstream benefits, including improve parameter initialization, better estimates of required network expansion, and better informed consolidation policies.
\end{enumerate}



\subsection{Convolutional Networks}
\label{sec:rd_cnns}

\begin{figure}[ht]
    \centering
    \includegraphics[width=0.8\textwidth]{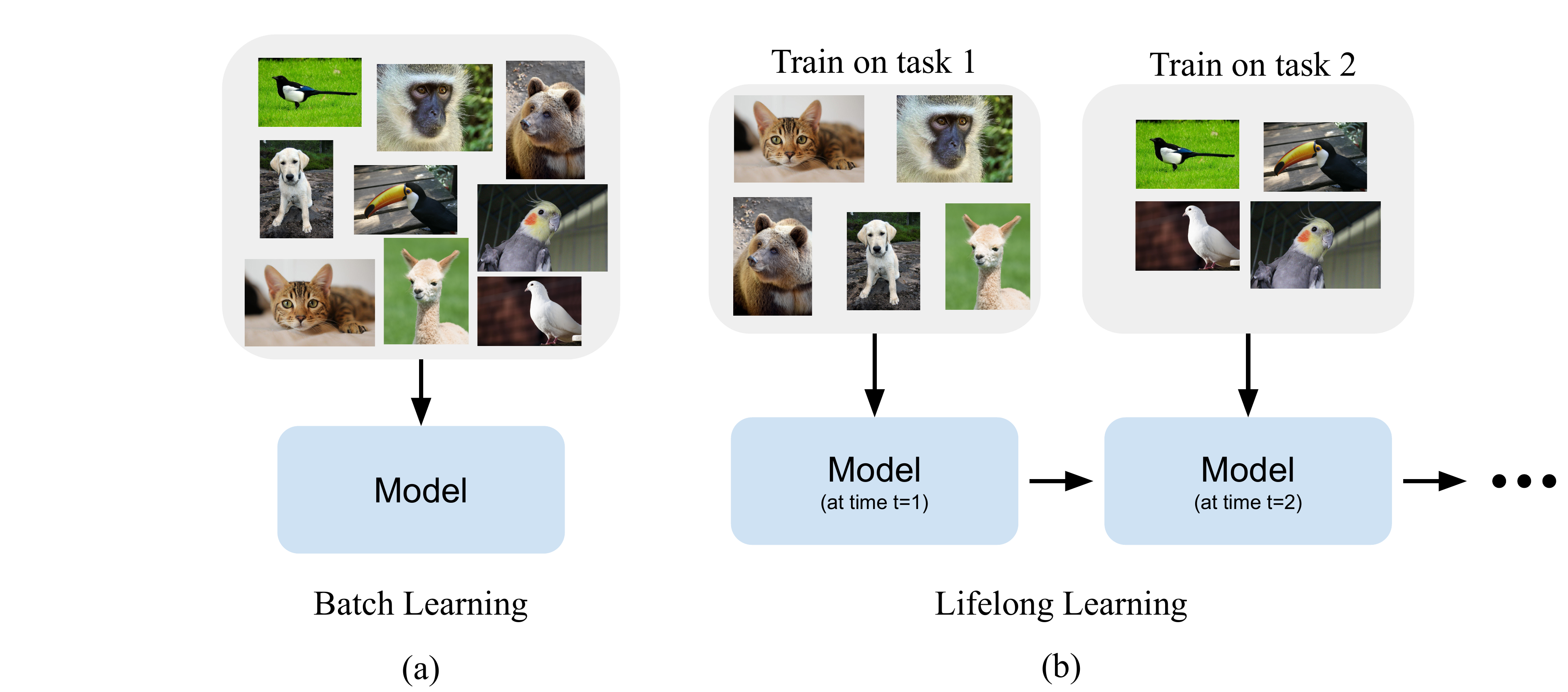}
    \caption{A comparison between batch learning generally commonly applied for computer vision, and the more human-like LLL approach. With batch learning, all data is combined and trained together in one sitting. With LLL, improved memory and computational efficiency can be attained by training on individual tasks (such as mammals in $T_{1}$ and then birds in $T_{2}$) in succession. This is done without requiring previous task data or forgetting of previous tasks, similar to humans.}
    \label{fig:batchvslll}
\end{figure}

While we have so far considered our framework as applied to fully connected feed-forward networks, we can also consider its application to convolutional neural networks (CNNs). Training on image tasks is often performed with resource-intensive batch learning, while our framework would allow for increased efficiency while being careful to maintain the high performance associated with batch learning. For example, our framework could first learn to classify mammals as $T_{1}$, and birds as $T_{2}$ (as shown in Figure~\ref{fig:batchvslll}). This is closer to human learning, as we never learn to recognize all objects in a single sitting -- which is how training commonly occurs on the popular ImageNet dataset \cite{Imagenet} for example. Instead, we are exposed to different things over the course of our lives. This would be done without re-using data for $T_{1}$ or forgetting $T_{1}$, while allowing forward transfer from $T_{1}$ to $T_{2}$ (ways of achieving these properties, among others, are discussed in Sec.~\ref{sec:framework}).

\begin{figure}[ht]
    \centering
    \includegraphics[width=0.8\textwidth]{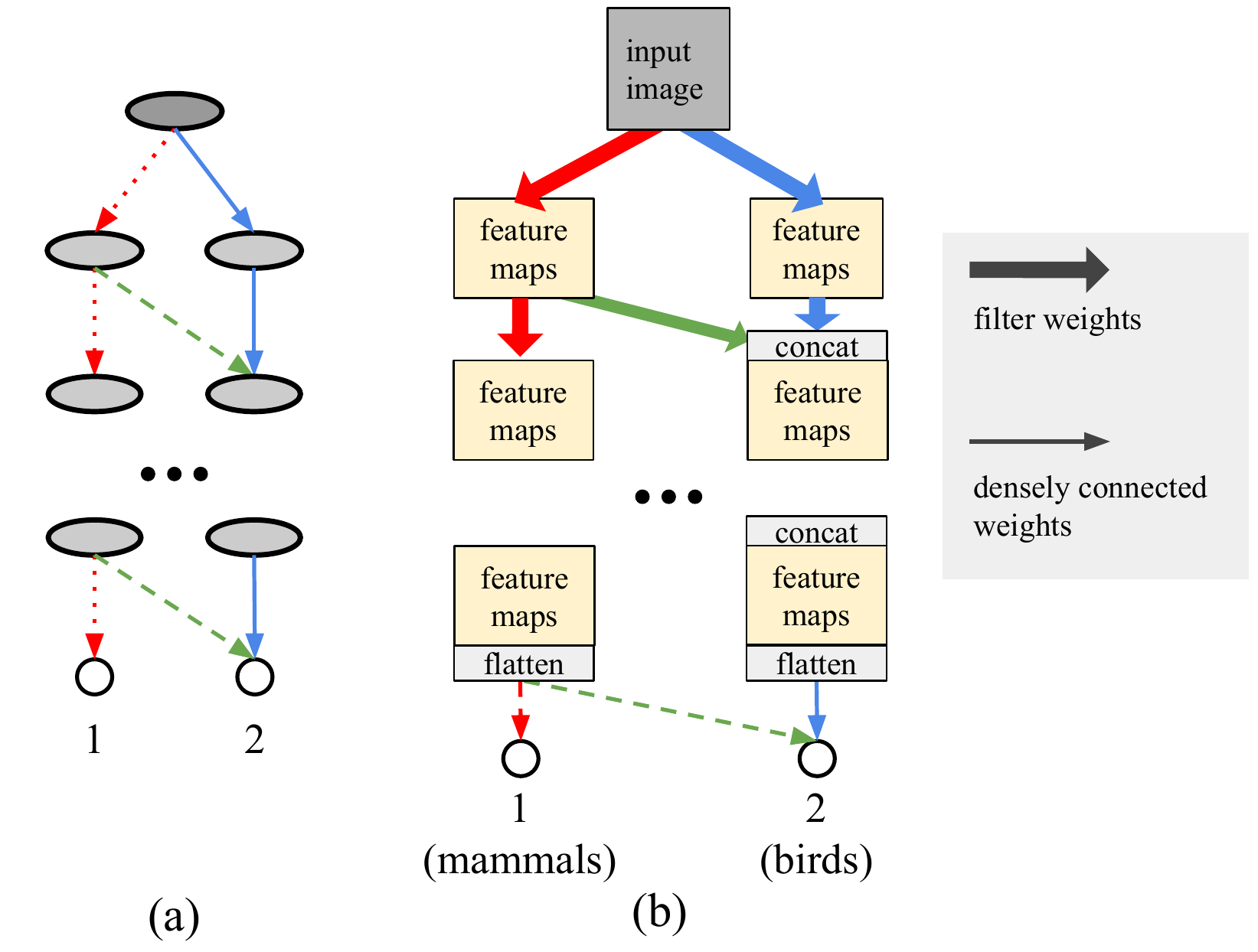}
    \caption{A simplified visualization of how our framework can be applied to convolutional networks. In (a) is a diagram of our consolidation mechanism and expansion being applied for learning two tasks with a fully connected network. In (b) is the corresponding CNN with the same high-level topology as (a) being applied to e.g. mammals for $T_{1}$ and birds for $T_{2}$. Thick links indicate filter weights, and thin links indicate weights between fully connected layers. Red indicates large $\pmb{b}$ values, blue indicated $\pmb{b}$ values of 0, and green indicates transfer links whose consolidation may depend on task similarity.}
    \label{fig:cnn}
\end{figure}

Adapting the consolidation mechanism to work with CNNs can be done as follows. In the fully connected networks, each consolidation value $\pmb{b}_{i}$ corresponds to a weight in a densely connected layer, but in CNNs, each $\pmb{b}_{i}$ corresponds to a filter. A filter is essentially a set of weights which takes the representation from the previous layer (feature maps), and transforms it into a new feature map. Large $\pmb{b}$ on a filter means the filter cannot be easily modified during learning. With the consolidation mechanism described for CNNs, we can now consider how to achieve the various LLL properties in CNNs:

\textbf{Continual learning of new tasks}. To extend the network for new tasks, we now add columns of convolutional filters, as reflected in Figure~\ref{fig:cnn} (blue links). Transfer links between tasks (thick green arrow in Fig.~\ref{fig:cnn}) now correspond to filters. The resultant feature maps of these ``transfer filters'' are concatenated with the feature maps in the new task column.  For more difficult tasks, we can add a greater number of filters (instead of a greater number of nodes). Instead of pruning weights or nodes, we can now prune filters.

\textbf{Non-forgetting}. Controlling non-forgetting means that large values in $\pmb{b}$ are applied to filters of previous tasks so they are not easily modifiable. See red links of Fig.~\ref{fig:cnn}.

\textbf{Forward transfer}. The two methods of encouraging forward transfer also work with CNNs. First, instead of ``copying'' weights for previous tasks when similar to the new task, we can now copy over filter values. This can be done using similar ideas to those laid out in Sec.~\ref{sec:forwardtransfer}. Second, we can encourage forward transfer and avoid negative transfer through controlling flexibility of the transfer filters (green links in Fig.~\ref{fig:cnn}). When these filters are initialized to zero and the $\pmb{b}$ values are very large, the new task can be prevented from using knowledge from the previous task.

While dynamic CNN architectures for lifelong learning have been considered before, starting from the AlexNet architecture \cite{yoon2018lifelong, krizhevsky2012imagenet}, we can consider taking a step further and considering more complex units like DenseNet blocks \cite{huang2017densely} or Inception modules \cite{szegedy2015going}. It is clear that almost all of the techniques discussed in this paper can be extended to CNNs and more complex networks allowing the desired LLL skills (such as non-forgetting, forward transfer, few-shot learning, and so on) to be obtained. However, much more future research on this would be needed. 


\subsection{Deep Random Networks}
\label{sec:rd_rand}

Another active research area that can be viewed from our LLL framework is deep random neural networks, which have been found to provide efficient learning with good generalization bounds \cite{rudi2017generalization, rosenfeld2019intriguing, gallicchio2020deep}. Their success derives from the surprising usefulness of random projections at tasks such as locality-sensitive hashing \cite{giryes2016deep, andoni2008near}.

The design of this lifelong learning approach, summarized in Algorithm~\ref{alg:rand_nn} is as follows: randomly initialize a neural network and permanently freeze all but the output layer (via the consolidation policy). To learn new tasks, only train the weights to the new output nodes. 

\begin{algorithm}[H]
    \SetAlgoLined
    Initialize a single random network \tcp{Gaussian weights have been proven to work well for this kind of approach \cite{giryes2016deep}}
    \textcolor{Cerulean}{$\CIRCLE$} Freeze all weights except for the output layer \tcp{red weights in Fig.~\ref{fig:rand_nn}}
    \textcolor{Cerulean}{$\CIRCLE$} Utilize Algorithms \ref{alg:forward_transfer}, \ref{alg:confusion_reduction}, \ref{alg:overall_refinement} during lifelong training on task sequence \tcp{for forward transfer, confusion reduction, and overall refinement respectively}
\caption{Learning using a Random Network}
\label{alg:rand_nn}
\end{algorithm}

This approach provides guaranteed non-forgetting. Other benefits include computational efficiency (only a minority of weights need to be tuned), and sample storage size efficiency (instead of storing whole samples, we can simply store the activation vectors at last hidden layer and skip a majority of inference during rehearsal).

\begin{figure}[ht]
    \centering
    \includegraphics[width=0.7\textwidth]{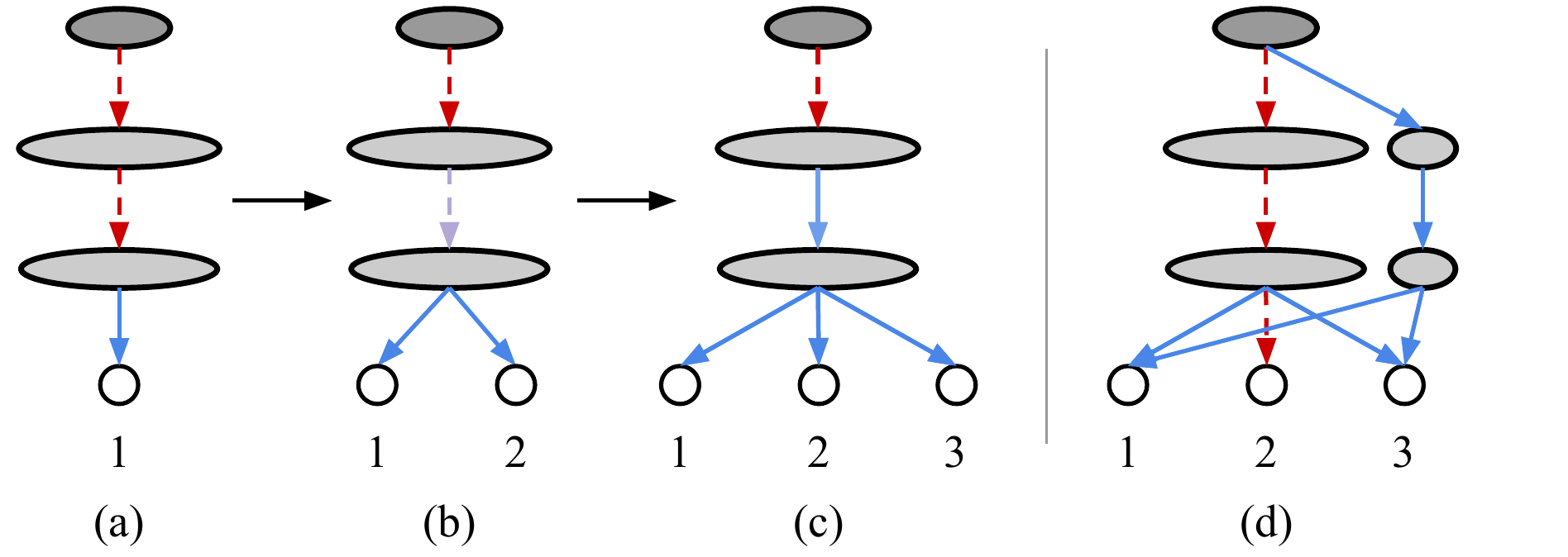}
    \caption{Visualization of the deep random projection lifelong learning approach with possible iterative unfreezing shown in (a) to (b) to (c). Initially, all weights except for the output layer are frozen. Slowly, layers (starting from the near the output) are unfrozen (transitioning from red to blue). In (d) is an example of how expansion with column-wise freezing and unfreezing can be used in the case where confusion (e.g. between $T_{1}$ and $T_{3}$) cannot be reduced with the original architecture. }
    \label{fig:rand_nn}
\end{figure}

With regard to this kind of approach, some possible research directions include:

\begin{enumerate}
    \item Unfreezing even just a small fraction of the weights was found to improve performance considerably \cite{rosenfeld2019intriguing}. If this is done, can we still maintain LLL skills that come with this approach such as guaranteed (or at least tightly bounded) non-forgetting and computational and storage efficiency?
    \item We can consider slowly unfreezing starting from the last hidden layer and moving towards the input as we learn more tasks, such as in Figure~\ref{fig:rand_nn} (a), (b), (c). Combined with overall refinement to prevent any forgetting of earlier tasks, this would allow the network to accommodate more difficult tasks while still maintaining some of the efficiency of a fully-frozen network.
    \item It would be an interesting research question to determine the optimal consolidation policy or schedule with which to progressively unfreeze layers. Should flexibility only be increased when we are unable to achieve sufficient performance on tasks?
    \item While weights can be frozen layer-wise, among unfrozen layers, we can still apply our LLL algorithms discussed earlier to freeze and unfreeze weights for new task learning without forgetting, forward- and backward-transfer, confusion reduction, and so on. That is, weights can be frozen and unfrozen layer-wise as well as task-wise according to the need. Fig.~\ref{fig:rand_nn}d for example illustrates how confusion reduction might be performed when the original randomly initialized network is insufficient to resolve confusion. After resolving confusion, the newly added weights can be frozen similar to the others in their layers.
\end{enumerate}

\subsection{Capsule Networks}
\label{sec:rd_capsules}

We believe that there is considerable overlap between capsule networks \cite{sabour2017dynamic, hinton2018matrix} and our LLL framework.  
Capsule networks emphasize modularity in neural network learning for computer vision; that is, capsules are a group of neurons that can accomplish certain CV tasks. In LLL, capsules are analogous to sub-networks (such as sub-networks for tasks 1, 2, 3 in Fig.~\ref{fig:framework_p1}) for different tasks. Dynamic routing in capsule networks is analogous to forward- and backward-transfer links that connect different sub-networks in LLL. The main difference, in our view, is that LLL emphasizes that tasks are often learned in sequence, usually from simple to complex, over time.  For humans, LLL happens over many years.  Usually easier shapes would be learned earlier in life directly with supervised data, and then modules would be frozen to prevent forgetting (Sec.~\ref{sec:nonforgetting}), and its outputs can be used to learn more complex shapes. 
Of course human learning is not always done from simple to complex.  Often data from an earlier task can arrive later and an already-learned learned module must be updated. This is discussed in Sec~\ref{sec:conceptdrift}, where the weights for all other tasks are frozen to prevent forgetting while knowledge for a task is being updated with new data.

\begin{figure}[ht]
    \centering
    \includegraphics[width=0.75\textwidth]{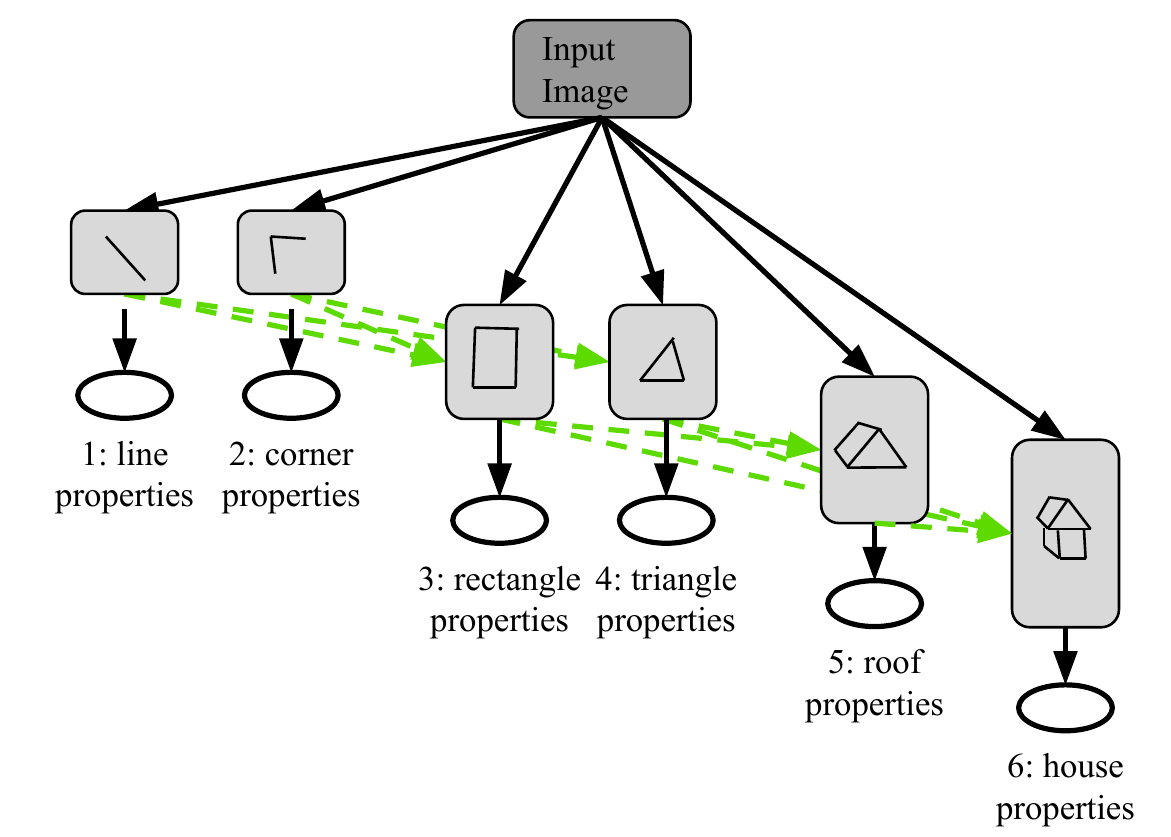}
    \caption{An illustrative example of learning multiple tasks of various complexity in LLL.  Often simpler tasks (such as lines) are learned earlier with training data (of lines).  The weights in their sub-networks or modules would be frozen to prevent forgetting, and their outputs would be forward-transferred to other modules to participate in learning other tasks (such as squares, roof, and house).  The weights between modules are also learned as in forward and backward transfer (Sec~\ref{sec:forwardtransfer}, \ref{sec:backwardtransfer}), regularized by the consolidation values.  Sub-networks or modules are like capsules and links are like dynamic routing in Capsule Networks. Ideally LLL happens from simple to complex, but often can be episodic (Sec.~\ref{sec:conceptdrift}).  
    In the figure, black links can be frozen or unfrozen based on the need. We also use the size of the module to roughly represent the complexity of the module, or the depths and number of nodes in the network capsules.  Note that here the ``modules'' can be neural networks, or other learning modules, as long as there is a consolidation parameter controlling how flexibly the module can be modified in learning.}
    \label{fig:capsule_like}
\end{figure}

Figure~\ref{fig:capsule_like} illustrates the LLL of various vision tasks in a modular fashion. 
Note that here the ``modules'' can be neural networks, or other learning algorithms, as long as there is a consolidation parameter controlling how flexibly the module can be modified in learning.
The whole LLL process is complex, and various steps have been discussed in Sec.~\ref{sec:framework} and the previous parts of this section. 

A real-life example of how metaphoric ``capsules'' in human learning can be illustrated as follows: 
The design of houses is different between cultures, climates, and geographic locations, and evolves over time. 
Seeing a straw roof for the first time may be surprising to many people. After seeing a straw roof, the module for ``roof'' may be re-learned (while not forgetting other learned concepts, such as walls) after checking with local residents that straw can in fact be used for a roof (supervised example).  
This new knowledge improves the robustness of recognizing houses, whether they have shingles or straw roofs, and can provide forward transfer when learning to recognize further types of houses and roofs.

%% file: sec_5_parallels.tex
\section{Speculations on Parallels with Human Learning}
\label{sec:parallels}

As our framework focuses on controlling the flexibility of individual network weights, it is natural to ask how our approach lines up with the mammalian brain whose lifelong concept learning behaviours we are attempting to capture.
In this section we first consider how human brains may have come to exhibit the kind of procedures discussed in Sec.~\ref{sec:framework}. Next, we explore parallels between our framework and human learning with respect to several behavioural patterns listed in Table~\ref{tab:human}. For each of these, we will discuss how a similar behaviour can be exhibited by our framework with specific network sizes and consolidation policies, 
and how they may translate to what happens in the human brain at a similar level of abstraction. 


\subsection{Reflecting on Nature vs. Nurture}

One may wonder how humans acquire such consolidation strategies as used in our LLL framework. This question is related to the long-standing debate on the influences of nature vs. nurture on human intelligence \cite{leahy1935nature, gazzaniga1992nature, kan2013nature}. Here we only reflect on the issue as it relates to our framework.  

We believe that some of the consolidation policies in Sec.~\ref{sec:framework} are a result of nature -- intrinsic to the physiology of the brain and a product of natural evolution. This would include those relating to network expansion and pruning (Sec.\ref{sec:expansion}), non-forgetting (Sec.\ref{sec:nonforgetting}), forward transfer (Sec.\ref{sec:forwardtransfer}), adapting to concept drift (Sec.\ref{sec:conceptdrift}), and graceful forgetting (Sec.\ref{sec:gracefulforgetting}). Some effects of nature on learning take place in the following ways: People are born with many more free neurons and synapses than required in adulthood \cite{ackerman1992discovering} and appear to have an innate capacity for learning many concepts (and other skills for the world around them) without catastrophic forgetting
(perhaps using algorithms similar to those described in Sec.~\ref{sec:framework} and \ref{sec:connections}). 
As we develop, neurons will be pruned without our conscious effort, especially starting around adolescence \cite{sisk2005pubertal}. Throughout adulthood, graceful forgetting, refinement of existing concepts, and adapting to slowly shifting concepts and environments naturally occurs \cite{ackerman1992discovering, mcewen2010stress}.

In contrast, learning specific concepts and knowledge from data are more likely to be the result of nurture -- they are things we must learn (such as recognizing different animals, shapes, and fruits). However, a person can perhaps consciously modify their nature-provided algorithms by performing meta-learning and changing consolidation policies. An example of this is memory training and explicit rehearsal.  This is  analogous to re-training of a specific concept with additional data while slowly increasing the $\pmb{b}$ values for the concept in our framework.

\subsection{Interpreting Human Learning Properties}

We observe that our LLL framework seems to be able to explain human LLL well, with many properties discussed below and summarized in Table~\ref{tab:human}. 
Here we broaden our scope to general human learning of tasks and skills, instead of restricted to supervised learning. 
Further research is of course needed to determine to what extent human brains actually use mechanisms similar to those discussed in this paper.

\textbf{Resourceful and versatile}. Ideally, human learning allows us to acquire a lot of knowledge, yet be flexible enough to adapt to new experiences and make meaningful connections between experiences.
This is analogous to our framework in which the network has unlimited free units for new tasks (Sec.~\ref{sec:expansion}), can utilize previous knowledge to learn new tasks (Sec.~\ref{sec:forwardtransfer}), will not likely forget previous tasks while learning new ones (Sec.~\ref{sec:nonforgetting}), and can use new knowledge to refine old skills (Sec.~\ref{sec:backwardtransfer}).

\textbf{Memory loss}.
In Section~\ref{sec:nonforgetting}, we describe how by setting values in $\pmb{b}$ for previous tasks to be large (freezing), previous tasks will not be easily forgotten. However, if not, ``memory loss'' can gradually happen.  This can also happen when the available free units are limited, so that graceful forgetting (Sec.~\ref{sec:gracefulforgetting}) must happen in order to learn new tasks. If the kind of pruning performed for graceful forgetting is too aggressive, further detrimental memory loss may occur.

\textbf{Sleep deprived}. The human brain is suspected of performing important memory-related processes during sleep, and sleep deprivation has been observed to be detrimental to cognitive performance related to memory \cite{walker2010sleep, killgore2010effects}. Confusion reduction and backward transfer are important stages of our proposed approach which utilize rehearsal (functionally similar to memory replay) -- where the model is exposed to samples from past tasks in order to perform fine-tuning to achieve various properties (Sec.~\ref{sec:confusionreduction} and \ref{sec:backwardtransfer}). Without these rehearsal-utilizing steps, the model may be less able to distinguish between samples of similar classes. Additionally, the ability to identify connections between newer tasks and older ones will be lost, so that potentially useful newly acquired skills cannot benefit older tasks. In addition, when the brain is ``tired'' and not well-rested, this may be analogous to poor optimization of policies in our framework (and in all machine learning algorithms). If optimization is not thorough in Equation~\ref{eqn:loss}, performance would be poor in all aspects of LLL. 

\textbf{``Rain man''}.
If values in $\pmb{b}$ for previous tasks (Sec.~\ref{sec:nonforgetting}) are set to very large to achieve non-forgetting, and no connections are made between previous and new tasks, the knowledge transfer (Sec.~\ref{sec:forwardtransfer} and \ref{sec:backwardtransfer}) will likely not happen.  
This is reminiscent of Kim Peek, who was able to remember vast amounts of information, but performed poorly at abstraction-related tasks that require connecting unrelated skills and information and was the inspiration for the main character of the movie \textit{Rain Man} \cite{treffert2005inside}.

\textbf{Alzheimer's disease}. Alzheimer's disease is a highly complex neurological disease and is not fully understood by the medical community. However, an early stage of Alzheimer's disease is usually characterized by good memory on events years ago but poor on recent events \cite{tierney1996prediction}. This can be modeled in our framework by a small number of new neurons for new tasks and a very large $\pmb{b}$ values for old ones.

\begin{table}[ht]
    \centering
    \resizebox{\textwidth}{!}{%
    \begin{tabular}{@{}llll@{}}
    \toprule
    \textbf{Behaviour} & \textbf{\# of neurons} & \textbf{$\pmb{b}$-setting policy} \\ \midrule
    \textbf{Resourceful and versatile} & large & large $\pmb{b}$ for non-forgetting (Sec.~\ref{sec:nonforgetting}), small $\pmb{b}$ for transfer (Sec.~\ref{sec:forwardtransfer}, \ref{sec:backwardtransfer}) \\
    \textbf{Memory loss} & small & small $\pmb{b}$ for previous tasks (Sec.~\ref{sec:nonforgetting}) and aggressive pruning  (Sec.~\ref{sec:gracefulforgetting})\\
    \textbf{Sleep deprived} & - & poor optimization of Eq.~\ref{eqn:loss} and application of consolidation polities\\
    \textbf{``Rain man''} & large & large $\pmb{b}$ for old tasks (Sec.~\ref{sec:nonforgetting}) and transfer links (Sec.~\ref{sec:forwardtransfer}, \ref{sec:backwardtransfer}) \\
    \textbf{Alzheimer’s disease} & small/shrinking & large $\pmb{b}$ for oldest memories (Sec.~\ref{sec:nonforgetting}) and aggressive pruning  (Sec.~\ref{sec:gracefulforgetting}) \\ \bottomrule
    \end{tabular}%
    }
    \caption{Correspondences between settings in our LLL framework and human behavioural descriptions. For example, being able to transfer knowledge across concepts and from old concepts to a newly encountered one allows us to be versatile.
    Here we broaden our scope to general human learning of tasks and skills, not restricting to supervised learning. }
    \label{tab:human}
\end{table}

%% file: sec_6_conclusion.tex
\section{Conclusions}
\label{sec:conclusions}

In this work, we surveyed many topics and previous works related to LLL and presented a conceptual unified framework for LLL using one central mechanism based on consolidation. 
First we discussed how our approach can capture many important properties of lifelong learning of concepts, including non-forgetting, forward and backward transfer, confusion reduction, and so on, under one roof. 
Second, we showed how our framework can be applied to many learning settings such as multi-task learning and few-shot learning.  
Rather than aiming for state-of-the-art results, this paper proposed research directions to help inform future lifelong learning research, including new algorithms and theoretical results. 
Last, we noted several similarities between our models with different consolidation policies and certain behaviors in human learning. 

%% file: main.bbl
\begin{thebibliography}{10}

\bibitem{litjens2017survey}
Geert Litjens, Thijs Kooi, Babak~Ehteshami Bejnordi, Arnaud Arindra~Adiyoso
  Setio, Francesco Ciompi, Mohsen Ghafoorian, Jeroen~Awm Van Der~Laak, Bram
  Van~Ginneken, and Clara~I S{\'a}nchez.
\newblock A survey on deep learning in medical image analysis.
\newblock {\em Medical image analysis}, 42:60--88, 2017.

\bibitem{shen2017deep}
Dinggang Shen, Guorong Wu, and Heung-Il Suk.
\newblock Deep learning in medical image analysis.
\newblock {\em Annual review of biomedical engineering}, 19:221--248, 2017.

\bibitem{radford2019language}
Alec Radford, Jeffrey Wu, Rewon Child, David Luan, Dario Amodei, and Ilya
  Sutskever.
\newblock Language models are unsupervised multitask learners.
\newblock {\em OpenAI Blog}, 1(8):9, 2019.

\bibitem{devlinetal2019bert}
Jacob Devlin, Ming-Wei Chang, Kenton Lee, and Kristina Toutanova.
\newblock {BERT}: Pre-training of deep bidirectional transformers for language
  understanding.
\newblock In {\em Proceedings of the 2019 Conference of the North {A}merican
  Chapter of the Association for Computational Linguistics: Human Language
  Technologies, Volume 1 (Long and Short Papers)}, pages 4171--4186,
  Minneapolis, Minnesota, June 2019. Association for Computational Linguistics.

\bibitem{silver2017mastering}
David Silver, Julian Schrittwieser, Karen Simonyan, Ioannis Antonoglou, Aja
  Huang, Arthur Guez, Thomas Hubert, Lucas Baker, Matthew Lai, Adrian Bolton,
  et~al.
\newblock Mastering the game of go without human knowledge.
\newblock {\em Nature}, 550(7676):354--359, 2017.

\bibitem{kamilaris2018deep}
Andreas Kamilaris and Francesc~X Prenafeta-Bold{\'u}.
\newblock Deep learning in agriculture: A survey.
\newblock {\em Computers and electronics in agriculture}, 147:70--90, 2018.

\bibitem{caruana1997multitask}
Rich Caruana.
\newblock Multitask learning.
\newblock {\em Machine learning}, 28(1):41--75, 1997.

\bibitem{pan2009survey}
Sinno~Jialin Pan and Qiang Yang.
\newblock A survey on transfer learning.
\newblock {\em IEEE Transactions on knowledge and data engineering},
  22(10):1345--1359, 2009.

\bibitem{fei2006one}
Li~Fei-Fei, Rob Fergus, and Pietro Perona.
\newblock One-shot learning of object categories.
\newblock {\em IEEE transactions on pattern analysis and machine intelligence},
  28(4):594--611, 2006.

\bibitem{thrun1998lifelong}
Sebastian Thrun.
\newblock Lifelong learning algorithms.
\newblock In {\em Learning to learn}, pages 181--209. Springer, 1998.

\bibitem{LLL}
S.~Thrun.
\newblock A lifelong learning perspective for mobile robot control.
\newblock {\em Intelligent Robots and Systems}, 1995.

\bibitem{parisi2019continual}
German~I Parisi, Ronald Kemker, Jose~L Part, Christopher Kanan, and Stefan
  Wermter.
\newblock Continual lifelong learning with neural networks: A review.
\newblock {\em Neural Networks}, 2019.

\bibitem{CatastrophicForgetting}
Michael McCloskey and Neal~J Cohen.
\newblock Catastrophic interference in connectionist networks: The sequential
  learning problem.
\newblock {\em The psychology of learning and motivation}, pages 109--165,
  1989.

\bibitem{de2019continual}
Matthias De~Lange, Rahaf Aljundi, Marc Masana, Sarah Parisot, Xu~Jia, Ales
  Leonardis, Gregory Slabaugh, and Tinne Tuytelaars.
\newblock Continual learning: A comparative study on how to defy forgetting in
  classification tasks.
\newblock {\em arXiv preprint arXiv:1909.08383}, 2019.

\bibitem{lake2015human}
Brenden~M Lake, Ruslan Salakhutdinov, and Joshua~B Tenenbaum.
\newblock Human-level concept learning through probabilistic program induction.
\newblock {\em Science}, 350(6266):1332--1338, 2015.

\bibitem{foundalis2006phaeaco}
Harry~E Foundalis.
\newblock Phaeaco: A cognitive architecture inspired by bongard's problems.
\newblock 2006.

\bibitem{bongard1967problem}
Mikhail~M Bongard.
\newblock The problem of recognition.
\newblock {\em Fizmatgiz, Moscow}, 1967.

\bibitem{yun2020deeper}
Xinyu Yun, Tanner Bohn, and Charles Ling.
\newblock A deeper look at bongard problems.
\newblock In {\em Canadian Conference on Artificial Intelligence}, pages
  528--539. Springer, 2020.

\bibitem{aljundi2018memory}
Rahaf Aljundi, Francesca Babiloni, Mohamed Elhoseiny, Marcus Rohrbach, and
  Tinne Tuytelaars.
\newblock Memory aware synapses: Learning what (not) to forget.
\newblock In {\em Proceedings of the European Conference on Computer Vision
  (ECCV)}, pages 139--154, 2018.

\bibitem{rebuffi2017icarl}
Sylvestre-Alvise Rebuffi, Alexander Kolesnikov, Georg Sperl, and Christoph~H.
  Lampert.
\newblock icarl: Incremental classifier and representation learning.
\newblock In {\em CVPR}, pages 5533--5542. IEEE Computer Society, 2017.

\bibitem{isele2018selective}
David Isele and Akansel Cosgun.
\newblock Selective experience replay for lifelong learning.
\newblock In {\em Thirty-second AAAI conference on artificial intelligence},
  2018.

\bibitem{chaudhry2019continual}
Arslan Chaudhry, Marcus Rohrbach, Mohamed Elhoseiny, Thalaiyasingam Ajanthan,
  Puneet~K Dokania, Philip~HS Torr, and Marc'Aurelio Ranzato.
\newblock Continual learning with tiny episodic memories.
\newblock {\em arXiv preprint arXiv:1902.10486}, 2019.

\bibitem{wu2019large}
Yue Wu, Yinpeng Chen, Lijuan Wang, Yuancheng Ye, Zicheng Liu, Yandong Guo, and
  Yun Fu.
\newblock Large scale incremental learning, 2019.

\bibitem{kirkpatrick2016EWC}
James Kirkpatrick, Razvan Pascanu, Neil Rabinowitz, Joel Veness, Guillaume
  Desjardins, Andrei~A. Rusu, Kieran Milan, John Quan, Tiago Ramalho, Agnieszka
  Grabska-Barwinska, Demis Hassabis, Claudia Clopath, Dharshan Kumaran, and
  Raia Hadsell.
\newblock Overcoming catastrophic forgetting in neural networks, 2016.
\newblock cite arxiv:1612.00796.

\bibitem{zenke2017synaptic}
Friedemann Zenke, Ben Poole, and Surya Ganguli.
\newblock Continual learning through synaptic intelligence.
\newblock In Doina Precup and Yee~Whye Teh, editors, {\em ICML}, volume~70 of
  {\em Proceedings of Machine Learning Research}, pages 3987--3995. PMLR, 2017.

\bibitem{chaudhry2018riemann}
Arslan Chaudhry, Puneet~Kumar Dokania, Thalaiyasingam Ajanthan, and Philip
  H.~S. Torr.
\newblock Riemannian walk for incremental learning: Understanding forgetting
  and intransigence.
\newblock {\em CoRR}, abs/1801.10112, 2018.

\bibitem{ritter2018online}
Hippolyt Ritter, Aleksandar Botev, and David Barber.
\newblock Online structured laplace approximations for overcoming catastrophic
  forgetting.
\newblock In {\em Advances in Neural Information Processing Systems}, pages
  3738--3748, 2018.

\bibitem{li2017learning}
Zhizhong Li and Derek Hoiem.
\newblock Learning without forgetting.
\newblock {\em IEEE transactions on pattern analysis and machine intelligence},
  40(12):2935--2947, 2017.

\bibitem{zhang2020class}
Junting Zhang, Jie Zhang, Shalini Ghosh, Dawei Li, Serafettin Tasci, Larry
  Heck, Heming Zhang, and C-C~Jay Kuo.
\newblock Class-incremental learning via deep model consolidation.
\newblock In {\em The IEEE Winter Conference on Applications of Computer
  Vision}, pages 1131--1140, 2020.

\bibitem{rusu2016progressive}
Andrei~A. Rusu, Neil~C. Rabinowitz, Guillaume Desjardins, Hubert Soyer, James
  Kirkpatrick, Koray Kavukcuoglu, Razvan Pascanu, and Raia Hadsell.
\newblock Progressive neural networks.
\newblock {\em CoRR}, abs/1606.04671, 2016.

\bibitem{yoon2018lifelong}
Jaehong Yoon, Eunho Yang, Jeongtae Lee, and Sung~Ju Hwang.
\newblock Lifelong learning with dynamically expandable networks.
\newblock In {\em ICLR (Poster)}. OpenReview.net, 2018.

\bibitem{xu2018reinforced}
Ju~Xu and Zhanxing Zhu.
\newblock Reinforced continual learning.
\newblock In {\em Advances in Neural Information Processing Systems}, pages
  899--908, 2018.

\bibitem{kading2016fine}
Christoph K{\"a}ding, Erik Rodner, Alexander Freytag, and Joachim Denzler.
\newblock Fine-tuning deep neural networks in continuous learning scenarios.
\newblock In {\em Asian Conference on Computer Vision}, pages 588--605.
  Springer, 2016.

\bibitem{ostapenko2019learning}
Oleksiy Ostapenko, Mihai Puscas, Tassilo Klein, Patrick Jahnichen, and Moin
  Nabi.
\newblock Learning to remember: A synaptic plasticity driven framework for
  continual learning.
\newblock In {\em Proceedings of the IEEE Conference on Computer Vision and
  Pattern Recognition}, pages 11321--11329, 2019.

\bibitem{li2019learn}
Xilai Li, Yingbo Zhou, Tianfu Wu, Richard Socher, and Caiming Xiong.
\newblock Learn to grow: A continual structure learning framework for
  overcoming catastrophic forgetting.
\newblock {\em arXiv preprint arXiv:1904.00310}, 2019.

\bibitem{golkar2019continual}
Siavash Golkar, Michael Kagan, and Kyunghyun Cho.
\newblock Continual learning via neural pruning.
\newblock {\em arXiv preprint arXiv:1903.04476}, 2019.

\bibitem{frankle2018lottery}
Jonathan Frankle and Michael Carbin.
\newblock The lottery ticket hypothesis: Finding sparse, trainable neural
  networks.
\newblock {\em arXiv preprint arXiv:1803.03635}, 2018.

\bibitem{zhang2016understanding}
Chiyuan Zhang, Samy Bengio, Moritz Hardt, Benjamin Recht, and Oriol Vinyals.
\newblock Understanding deep learning requires rethinking generalization.
\newblock {\em arXiv preprint arXiv:1611.03530}, 2016.

\bibitem{vuorio2018meta}
Risto Vuorio, Dong-Yeon Cho, Daejoong Kim, and Jiwon Kim.
\newblock Meta continual learning.
\newblock {\em arXiv preprint arXiv:1806.06928}, 2018.

\bibitem{zhang2017survey}
Yu~Zhang and Qiang Yang.
\newblock A survey on multi-task learning.
\newblock {\em arXiv preprint arXiv:1707.08114}, 2017.

\bibitem{kumar2012learning}
Abhishek Kumar and Hal Daume~III.
\newblock Learning task grouping and overlap in multi-task learning.
\newblock {\em arXiv preprint arXiv:1206.6417}, 2012.

\bibitem{ruvolo2013ella}
Paul Ruvolo and Eric Eaton.
\newblock Ella: An efficient lifelong learning algorithm.
\newblock In {\em International Conference on Machine Learning}, pages
  507--515, 2013.

\bibitem{wang2019characterizing}
Zirui Wang, Zihang Dai, Barnab{\'a}s P{\'o}czos, and Jaime Carbonell.
\newblock Characterizing and avoiding negative transfer.
\newblock In {\em Proceedings of the IEEE Conference on Computer Vision and
  Pattern Recognition}, pages 11293--11302, 2019.

\bibitem{morcos2019one}
Ari Morcos, Haonan Yu, Michela Paganini, and Yuandong Tian.
\newblock One ticket to win them all: generalizing lottery ticket
  initializations across datasets and optimizers.
\newblock In {\em Advances in Neural Information Processing Systems}, pages
  4933--4943, 2019.

\bibitem{chakraborty2015brain}
Mukta Chakraborty and Erich~D Jarvis.
\newblock Brain evolution by brain pathway duplication.
\newblock {\em Philosophical Transactions of the Royal Society B: Biological
  Sciences}, 370(1684):20150056, 2015.

\bibitem{oh2017zero}
Junhyuk Oh, Satinder Singh, Honglak Lee, and Pushmeet Kohli.
\newblock Zero-shot task generalization with multi-task deep reinforcement
  learning.
\newblock In {\em Proceedings of the 34th International Conference on Machine
  Learning-Volume 70}, pages 2661--2670. JMLR. org, 2017.

\bibitem{DEramo2020Sharing}
Carlo D'Eramo, Davide Tateo, Andrea Bonarini, Marcello Restelli, and Jan
  Peters.
\newblock Sharing knowledge in multi-task deep reinforcement learning.
\newblock In {\em International Conference on Learning Representations}, 2020.

\bibitem{smith2017federated}
Virginia Smith, Chao-Kai Chiang, Maziar Sanjabi, and Ameet~S Talwalkar.
\newblock Federated multi-task learning.
\newblock In {\em Advances in Neural Information Processing Systems}, pages
  4424--4434, 2017.

\bibitem{robins1995catastrophic}
Anthony Robins.
\newblock Catastrophic forgetting, rehearsal and pseudorehearsal.
\newblock {\em Connection Science}, 7(2):123--146, 1995.

\bibitem{shin2017continual}
Hanul Shin, Jung~Kwon Lee, Jaehong Kim, and Jiwon Kim.
\newblock Continual learning with deep generative replay.
\newblock In {\em Advances in Neural Information Processing Systems}, pages
  2990--2999, 2017.

\bibitem{atkinson2018pseudo}
Craig Atkinson, Brendan McCane, Lech Szymanski, and Anthony Robins.
\newblock Pseudo-recursal: Solving the catastrophic forgetting problem in deep
  neural networks.
\newblock {\em arXiv preprint arXiv:1802.03875}, 2018.

\bibitem{lopez2017gem}
David Lopez-Paz and Marc'Aurelio Ranzato.
\newblock Gradient episodic memory for continual learning.
\newblock In Isabelle Guyon, Ulrike von Luxburg, Samy Bengio, Hanna~M. Wallach,
  Rob Fergus, S.~V.~N. Vishwanathan, and Roman Garnett, editors, {\em NIPS},
  pages 6467--6476, 2017.

\bibitem{riemer2018learning}
Matthew Riemer, Ignacio Cases, Robert Ajemian, Miao Liu, Irina Rish, Yuhai Tu,
  and Gerald Tesauro.
\newblock Learning to learn without forgetting by maximizing transfer and
  minimizing interference.
\newblock {\em arXiv preprint arXiv:1810.11910}, 2018.

\bibitem{aljundi2019task}
Rahaf Aljundi, Klaas Kelchtermans, and Tinne Tuytelaars.
\newblock Task-free continual learning.
\newblock In {\em Proceedings of the IEEE Conference on Computer Vision and
  Pattern Recognition}, pages 11254--11263, 2019.

\bibitem{webb2016characterizing}
Geoffrey~I Webb, Roy Hyde, Hong Cao, Hai~Long Nguyen, and Francois Petitjean.
\newblock Characterizing concept drift.
\newblock {\em Data Mining and Knowledge Discovery}, 30(4):964--994, 2016.

\bibitem{bengio2009curriculum}
Yoshua Bengio, J{\'e}r{\^o}me Louradour, Ronan Collobert, and Jason Weston.
\newblock Curriculum learning.
\newblock In {\em Proceedings of the 26th annual international conference on
  machine learning}, pages 41--48. ACM, 2009.

\bibitem{elman1993learning}
Jeffrey~L Elman.
\newblock Learning and development in neural networks: The importance of
  starting small.
\newblock {\em Cognition}, 48(1):71--99, 1993.

\bibitem{ruvolo2013active}
Paul Ruvolo and Eric Eaton.
\newblock Active task selection for lifelong machine learning.
\newblock In {\em Twenty-seventh AAAI conference on artificial intelligence},
  2013.

\bibitem{sun2018active}
Gan Sun, Yang Cong, and Xiaowei Xu.
\newblock Active lifelong learning with" watchdog".
\newblock In {\em Thirty-Second AAAI Conference on Artificial Intelligence},
  2018.

\bibitem{girshick2014rich}
Ross Girshick, Jeff Donahue, Trevor Darrell, and Jitendra Malik.
\newblock Rich feature hierarchies for accurate object detection and semantic
  segmentation.
\newblock In {\em Proceedings of the IEEE conference on computer vision and
  pattern recognition}, pages 580--587, 2014.

\bibitem{wang2015unsupervised}
Xiaolong Wang and Abhinav Gupta.
\newblock Unsupervised learning of visual representations using videos.
\newblock In {\em Proceedings of the IEEE International Conference on Computer
  Vision}, pages 2794--2802, 2015.

\bibitem{mahajan2018exploring}
Dhruv Mahajan, Ross Girshick, Vignesh Ramanathan, Kaiming He, Manohar Paluri,
  Yixuan Li, Ashwin Bharambe, and Laurens van~der Maaten.
\newblock Exploring the limits of weakly supervised pretraining.
\newblock In {\em Proceedings of the European Conference on Computer Vision
  (ECCV)}, pages 181--196, 2018.

\bibitem{russakovsky2015imagenet}
Olga Russakovsky, Jia Deng, Hao Su, Jonathan Krause, Sanjeev Satheesh, Sean Ma,
  Zhiheng Huang, Andrej Karpathy, Aditya Khosla, Michael Bernstein, et~al.
\newblock Imagenet large scale visual recognition challenge.
\newblock {\em International journal of computer vision}, 115(3):211--252,
  2015.

\bibitem{simonyan2014very}
Karen Simonyan and Andrew Zisserman.
\newblock Very deep convolutional networks for large-scale image recognition.
\newblock {\em arXiv preprint arXiv:1409.1556}, 2014.

\bibitem{ramalho2019empirical}
Tiago Ramalho, Thierry Sousbie, and Stefano Peluchetti.
\newblock An empirical study of pretrained representations for few-shot
  classification.
\newblock {\em arXiv preprint arXiv:1910.01319}, 2019.

\bibitem{conneau2017supervised}
Alexis Conneau, Douwe Kiela, Holger Schwenk, Lo{\"\i}c Barrault, and Antoine
  Bordes.
\newblock Supervised learning of universal sentence representations from
  natural language inference data.
\newblock In {\em Proceedings of the 2017 Conference on Empirical Methods in
  Natural Language Processing}, pages 670--680, Copenhagen, Denmark, September
  2017. Association for Computational Linguistics.

\bibitem{peters2018deep}
Matthew~E Peters, Mark Neumann, Mohit Iyyer, Matt Gardner, Christopher Clark,
  Kenton Lee, and Luke Zettlemoyer.
\newblock Deep contextualized word representations.
\newblock {\em arXiv preprint arXiv:1802.05365}, 2018.

\bibitem{erhan2010does}
Dumitru Erhan, Yoshua Bengio, Aaron Courville, Pierre-Antoine Manzagol, Pascal
  Vincent, and Samy Bengio.
\newblock Why does unsupervised pre-training help deep learning?
\newblock {\em Journal of Machine Learning Research}, 11(Feb):625--660, 2010.

\bibitem{hospedales2020meta}
Timothy Hospedales, Antreas Antoniou, Paul Micaelli, and Amos Storkey.
\newblock Meta-learning in neural networks: A survey.
\newblock {\em arXiv preprint arXiv:2004.05439}, 2020.

\bibitem{Imagenet}
Ilya~Sutskever Alex~Krizhevsky and Geoffrey~E Hinton.
\newblock Imagenet classification with deep convolutional neural networks.
\newblock {\em NIPS}, page 1097–1105, 2012.

\bibitem{krizhevsky2012imagenet}
Alex Krizhevsky, Ilya Sutskever, and Geoffrey~E Hinton.
\newblock Imagenet classification with deep convolutional neural networks.
\newblock In {\em Advances in neural information processing systems}, pages
  1097--1105, 2012.

\bibitem{huang2017densely}
Gao Huang, Zhuang Liu, Laurens Van Der~Maaten, and Kilian~Q Weinberger.
\newblock Densely connected convolutional networks.
\newblock In {\em Proceedings of the IEEE conference on computer vision and
  pattern recognition}, pages 4700--4708, 2017.

\bibitem{szegedy2015going}
Christian Szegedy, Wei Liu, Yangqing Jia, Pierre Sermanet, Scott Reed, Dragomir
  Anguelov, Dumitru Erhan, Vincent Vanhoucke, and Andrew Rabinovich.
\newblock Going deeper with convolutions.
\newblock In {\em Proceedings of the IEEE conference on computer vision and
  pattern recognition}, pages 1--9, 2015.

\bibitem{rudi2017generalization}
Alessandro Rudi and Lorenzo Rosasco.
\newblock Generalization properties of learning with random features.
\newblock In {\em Advances in Neural Information Processing Systems}, pages
  3215--3225, 2017.

\bibitem{rosenfeld2019intriguing}
Amir Rosenfeld and John~K Tsotsos.
\newblock Intriguing properties of randomly weighted networks: Generalizing
  while learning next to nothing.
\newblock In {\em 2019 16th Conference on Computer and Robot Vision (CRV)},
  pages 9--16. IEEE, 2019.

\bibitem{gallicchio2020deep}
Claudio Gallicchio and Simone Scardapane.
\newblock Deep randomized neural networks.
\newblock In {\em Recent Trends in Learning From Data}, pages 43--68. Springer,
  2020.

\bibitem{giryes2016deep}
Raja Giryes, Guillermo Sapiro, and Alex~M Bronstein.
\newblock Deep neural networks with random gaussian weights: A universal
  classification strategy?
\newblock {\em IEEE Transactions on Signal Processing}, 64(13):3444--3457,
  2016.

\bibitem{andoni2008near}
Alexandr Andoni and Piotr Indyk.
\newblock Near-optimal hashing algorithms for near neighbor problem in high
  dimension.
\newblock {\em Communications of the ACM}, 51(1):117--122, 2008.

\bibitem{sabour2017dynamic}
Sara Sabour, Nicholas Frosst, and Geoffrey~E Hinton.
\newblock Dynamic routing between capsules.
\newblock In {\em Advances in neural information processing systems}, pages
  3856--3866, 2017.

\bibitem{hinton2018matrix}
Geoffrey~E Hinton, Sara Sabour, and Nicholas Frosst.
\newblock Matrix capsules with em routing.
\newblock 2018.

\bibitem{leahy1935nature}
Alice~M Leahy.
\newblock Nature-nurture and intelligence.
\newblock {\em Genetic Psychology Monographs}, 1935.

\bibitem{gazzaniga1992nature}
Michael~S Gazzaniga.
\newblock {\em Nature's mind: The biological roots of thinking, emotions,
  sexuality, language, and intelligence.}
\newblock Basic Books, 1992.

\bibitem{kan2013nature}
Kees-Jan Kan, Jelte~M Wicherts, Conor~V Dolan, and Han~LJ van~der Maas.
\newblock On the nature and nurture of intelligence and specific cognitive
  abilities: The more heritable, the more culture dependent.
\newblock {\em Psychological Science}, 24(12):2420--2428, 2013.

\bibitem{ackerman1992discovering}
Sandra Ackerman et~al.
\newblock {\em Discovering the brain}.
\newblock National Academies Press, 1992.

\bibitem{sisk2005pubertal}
Cheryl~L Sisk and Julia~L Zehr.
\newblock Pubertal hormones organize the adolescent brain and behavior.
\newblock {\em Frontiers in neuroendocrinology}, 26(3-4):163--174, 2005.

\bibitem{mcewen2010stress}
Bruce~S McEwen.
\newblock Stress, sex and neural adaptation to a changing environment:
  mechanisms of neuronal remodeling.
\newblock {\em Annals of the New York Academy of Sciences}, 1204(Suppl):E38,
  2010.

\bibitem{walker2010sleep}
Matthew~P Walker.
\newblock Sleep, memory and emotion.
\newblock In {\em Progress in brain research}, volume 185, pages 49--68.
  Elsevier, 2010.

\bibitem{killgore2010effects}
William~DS Killgore.
\newblock Effects of sleep deprivation on cognition.
\newblock In {\em Progress in brain research}, volume 185, pages 105--129.
  Elsevier, 2010.

\bibitem{treffert2005inside}
Darold~A Treffert and Daniel~D Christensen.
\newblock Inside the mind of a savant.
\newblock {\em Scientific American}, 293(6):108--113, 2005.

\bibitem{tierney1996prediction}
MC~Tierney, JP~Szalai, WG~Snow, RH~Fisher, A~Nores, G~Nadon, E~Dunn, and PH~St
  George-Hyslop.
\newblock Prediction of probable alzheimer's disease in memory-impaired
  patients: A prospective longitudinal study.
\newblock {\em Neurology}, 46(3):661--665, 1996.

\end{thebibliography}
